\documentclass[letter, 10 pt, conference]{ieeeconf}  

\IEEEoverridecommandlockouts                              

\usepackage[top=2.12cm, bottom=1.52cm, left=1.69cm, right=1.69cm]{geometry}
\let\labelindent\relax
\usepackage{enumitem}
\usepackage{graphics} 
\graphicspath{{Pictures/}}
\usepackage{epsfig} 
\usepackage{mathptmx} 
\usepackage{times} 
\usepackage{amsmath} 
\usepackage{amssymb}  
\usepackage{amsfonts}
\usepackage{graphicx}
\usepackage{tabularx}
\usepackage{multicol}
\usepackage{algorithm}
\usepackage{algorithmic}
\usepackage{multirow}
\usepackage{todonotes}%
\usepackage{lettrine}
\usepackage{wasysym}
\usepackage{subcaption}
\usepackage{bm}

\usepackage[font={scriptsize,it}]{caption} 
\usepackage[linewidth=1pt]{mdframed}
\usepackage{setspace}

\newcommand{\argmin}[1]{\underset{#1}{\operatorname{\mathbf{arg}}\,\operatorname{\mathbf{min}}}\;}
\newcommand{\x}{\mathbf{x}}
\newcommand{\xproj}{\mathbf{\check{x}}}

\newcommand{\ao}{\mathbf{u}}

\usepackage{cite}
\usepackage[bookmarks=false]{hyperref}
\usepackage{leftidx}

\definecolor{lightlightgrey}{rgb}{0.9,0.9,0.9}
\definecolor{Red}{rgb}{1,0,0}
\definecolor{Blue}{rgb}{0,0,1}
\definecolor{Green}{rgb}{0,1,0}
\definecolor{magenta}{rgb}{1,0,.6}
\definecolor{lightblue}{rgb}{0,.5,1}
\definecolor{lightpurple}{rgb}{.6,.4,1}
\definecolor{gold}{rgb}{.6,.5,0}
\definecolor{orange}{rgb}{1,0.4,0}
\definecolor{hotpink}{rgb}{1,0,0.5}
\definecolor{newcolor2}{rgb}{.5,.3,.5}
\definecolor{newcolor}{rgb}{0,.3,1}
\definecolor{newcolor3}{rgb}{1,1,1}
\definecolor{darkgreen1}{rgb}{0, .35, 0}
\definecolor{darkgreen}{rgb}{0, .6, 0}
\definecolor{darkred}{rgb}{.75,0,0}

\xdefinecolor{olive}{cmyk}{0.64,0,0.95,0.4}
\xdefinecolor{purpleish}{cmyk}{0.75,0.75,0,0}

\makeatletter
\IEEEtriggercmd{\reset@font\normalfont\fontsize{7.9pt}{8.40pt}\selectfont}
\makeatother
\IEEEtriggeratref{1}

\xdefinecolor{red}{rgb}{1,0,0}
\usepackage{graphicx}
\usepackage{color}
\usepackage{float}
\floatstyle{boxed}
\newfloat{math}{thp}{lop}
\floatname{math}{Equation}

\makeatother

\begin{document}

\title{Decentralized MPC based Obstacle Avoidance \\ for Multi-Robot Target Tracking Scenarios}

\author{Rahul Tallamraju$^{1,3}$, Sujit Rajappa$^2$, Michael Black$^1$, Kamalakar Karlapalem$^3$ and Aamir Ahmad$^1$
\thanks{\tiny{rahul.tallamraju, black, aamir.ahmad@tuebingen.mpg.de, sujit.rajappa@uni-tuebingen.de, kamal@iiit.ac.in}}
\thanks{\tiny{$^1$Perceiving Systems Department, Max Planck Institute for Intelligent Systems, T\"ubingen, Germany.}}
\thanks{\tiny{$^2$The Chair of Cognitive Systems, Department of Computer Science, University of T\"ubingen, T\"ubingen, Germany.}}
\thanks{\tiny{$^3$Agents and Applied Robotics Group, International Institute for Information Technology, Hyderabad, India.}}}
\maketitle


\pagestyle{empty}
\begin{abstract}
In this work, we consider the problem of decentralized multi-robot target tracking and obstacle avoidance in dynamic environments. Each robot executes a local motion planning algorithm which is based on model predictive control (MPC). The planner is designed as a quadratic program, subject to constraints on robot dynamics and obstacle avoidance. Repulsive potential field functions are employed to avoid obstacles. The novelty of our approach lies in embedding these non-linear potential field functions as constraints within a convex optimization framework. Our method convexifies non-convex constraints and dependencies, by replacing them as pre-computed external input forces  in robot dynamics. The proposed algorithm additionally incorporates different methods to avoid field local minima problems associated with using potential field functions in planning. The motion planner does not enforce predefined trajectories or any formation geometry on the robots and is a comprehensive solution for cooperative obstacle avoidance in the context of multi-robot target tracking. Video of simulation studies: \url{https://youtu.be/umkdm82Tt0M}
\end{abstract}

\section{Introduction}

\begin{figure}
\centering
\includegraphics[scale=0.3]{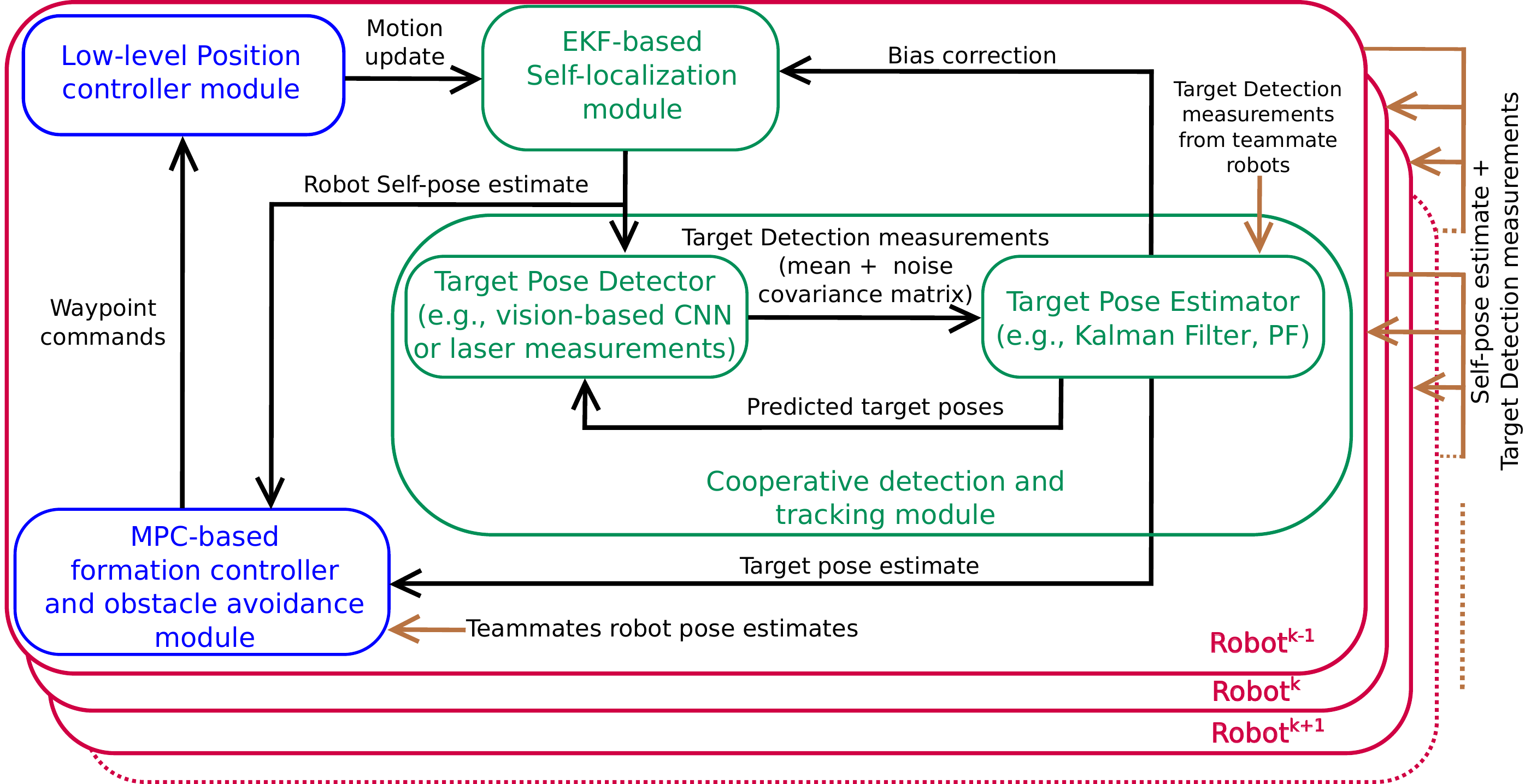}
\caption{General modules involved in multi-robot target tracking. Our work focuses on the modules highlighted in blue}
\label{system}
\vspace{-1.5em}
\end{figure}

The topic of multi-robot cooperative target tracking has been researched extensively in recent years \cite{aamir_pcmmc_ras_journal,dias2013cooperative,ahmad2017online,hausman2016cooperative,zhang2016cooperative}.  Target, here referes to a movable subject of interest in the environment for e.g., human, animal or other robot. Cooperative target tracking methods focus on improving the estimated pose of a tracked target while simultaneously enhancing the localization estimates of poorly localized robots, e.g., \cite{aamir_pcmmc_ras_journal}, by fusing the state estimate information acquired from team mate robots. The general modules involved in decentralized multi-robot target tracking are summarized in Fig. \ref{system}. Our work focuses on the modules related to obstacle avoidance (blue in Fig. \ref{system}). The other related modules (green in Fig. \ref{system}), such as target pose estimation, are assumed to be available (see \cite{price2018deep}). The developed obstacle avoidance module fits into any general cooperative target tracking framework as seen in Fig. \ref{system}. 

Robots involved in tracking a desired target must not collide with each other, and also with other entities (human, robot or environment). While addressing this problem, the state-of-art methodologies for obstacle avoidance in the context of cooperative target tracking have drawbacks. In \cite{nascimento2015nonlinear,nascimento2016multi} obstacle avoidance is imposed as part of the weighted MPC based optimization objective, thereby providing no guaranteed avoidance. In \cite{aamir_pcmmc_ras_journal} obstacle avoidance is a separate planning module, which modifies the generated optimization trajectory using potential fields. This leads to a sub-optimal trajectory and field local minima. 

The goal of this work is to provide a holistic solution to the problem of obstacle avoidance, in the context of multi-robot target tracking in an environment with static and dynamic obstacles. Our solution is an asynchronous and decentralized, model-predictive control based convex optimization framework. Instead of directly using repulsive potential field functions to avoid obstacle, we convexify the potential field forces by replacing them as pre-computed external input forces in robot dynamics. As long as a feasible solution exists for the optimization program, obstacle avoidance is guaranteed. In our proposed solution we present three methods to resolve the field local minima issue. This facilitates convergence to a desired surface around the target. 

The main contributions of this work are,
\begin{itemize}
\item Fully convex optimization for local motion planning in the context of multi-robot target tracking
\item Handling non-convex constraints as pre-computed input forces in robot dynamics, to enforce convexity
\item Guaranteed static and dynamic obstacle avoidance
\item Asynchronous, decentralized and scalable algorithm
\item Methodologies for potential field local minima avoidance
\end{itemize}

Sec.~\ref{sec:sota} details the state-of-art methods related to obstacle avoidance, Sec.~\ref{sec:proposedappraoch} discusses the Decentralized Quadratic Model Predictive Controller along with the proposed methodologies to solve the field local minima and control deadlock problem,  Sec.~\ref{sec:results} elaborates on simulation results of different scenarios, and finally we discuss future work directions. 

\section{State-of-the-Art}
\label{sec:sota} 

The main goal in this work is to develop a decentralized multi-robot target tracker and collision free motion planner in obstacle (static and dynamic) prone environments. The multi-agent obstacle avoidance problem has gained a lot of attention in recent years. Single agent obstacle avoidance, motion planning and control is well studied \cite{hoy2015algorithms,LIU2017317}. However, the multi-agent obstacle avoidance problem is more complex due to motion planning dependencies between different agents, and the poor computational scalability associated with the non-linear nature of these dependencies. 
In general, collision free trajectory generation for multi-agents can be classified into, (i) reactive and, (ii) optimization based approaches. 
Many reactive approaches are based on the concept of velocity obstacle (VO) \cite{fiorini1998motion}, whereas, optimization based approaches avoid obstacles by embedding collision constraints (like VO) within cost function or as hard constraints in optimization. 
Recently, a mixed integer quadratic program (MIQP) in the form of a centralized non-linear model predictive control (NMPC)\cite{fukushima2013model} has been proposed for dynamic obstacle avoidance, where feedback linearization is coupled with a variant of the branch and bound algorithm. 
However, this approach suffers with agent scale-up, since increase in binary variables of MIQP has an associated exponential complexity. 
In general, centralized optimization approaches \cite{raghunathan2004dynamic,kushleyev2013towards} are not computationally scalable with increase in number of agents. 
In \cite{derenick2007convex}, a decentralized convex optimization for multi-robot collision-free formation planning through static obstacles is studied.  This approach involves, triangular tessellation of the configuration space to convexify static obstacle avoidance constraints. Tessellated regions are used as nodes in a graph and the paths between the cells are determined to guarantee obstacle avoidance.
A decentralized NMPC \cite{shim2003decentralized} has been proposed for pursuit evasion and static obstacle avoidance with multiple aerial vehicles. Here the optimization is constrained by non-linear robot dynamics, resulting in non-convexity and thereby affecting the real-time NMPC performance. Additionally, a potential field function is also used as part of a weighted objective function for obstacle avoidance. A similar decentralized NMPC has been proposed for the task of multiple UAVs formation flight control \cite{chao2011collision}.  

Sequential convex programming (SCP) has been applied to solve the problems of multi-robot collision-free trajectory generation \cite{augugliaro2012generation}, trajectory optimization and target assignment \cite{morgan2015swarm} and formation payload transport \cite{alonso2015multi}. These methodologies principally approximate and convexify the non-convex obstacle avoidance constraints, and iteratively solve the resulting convex optimization problem until feasibility is attained. 
Due to this approximation, the obtained solutions are fast within a given time-horizon, albeit sub-optimal. 
SCPs have been very effective in generating real-time local motion plans with non-convex constraints. Recent work in multi-agent obstacle avoidance \cite{alonso2015collision} builds on the concept of reciprocal velocity obstacle \cite{van2011reciprocal}, where a local motion planner is proposed to characterize and optimally choose velocities that do not lead to a collision. 
The approach in \cite{alonso2015collision} convexifies the velocity obstacle (VO) constraint to guarantee local obstacle avoidance. 

In summary, due to non-linear dynamics constraints or obstacle avoidance dependencies, most multi-robot obstacle avoidance techniques are either, (i) centralized, (ii) non-convex, or (iii) locally optimal. Furthermore, some approaches only explore the solution space partially due to constraint linearization \cite{alonso2015collision}. Additionally, current NMPC target tracking approaches using potential fields do not provide guarantees on obstacle avoidance and are linked with the field local-minima problem. Recent reinforcement learning solutions \cite{chen2017decentralized, long2017deep} require large number of training scenarios to determine a policy and also do not guarantee obstacle avoidance.

Our work generates collision-free motion plans for each agent using a convex model-predicitive quadratic program in a decentralized manner. 
This approach guarantees obstacle avoidance and facilitates global convergence to a target surface. 
To the best of our knowledge, the method of using tangential potential field functions \cite{chang2003collision} to generate different reactive swarming behaviors including obstacle avoidance, is  most similar to our approach. However in \cite{chang2003collision}, the field local minima is persistent in the swarming behaviors. 

Unlike previous NMPC based target tracking approaches, which use potential fields in the objective, here  we use potential field forces as constraints in optimization. 
Specifically, the non-linear potential field functions are not directly used as constraints in optimization. Instead, potential field forces are pre-computed for a horizon using the horizon motion trajectory of neighboring agents and obstacles in the vicinity.  A feasible solution of the optimization program guarantees obstacle avoidance.  The pre-computed values are applied as external control input forces in the optimization process thereby preserving the overall convexity. 

%
\section{Proposed Approach}
\label{sec:proposedappraoch} 
\subsection{Preliminaries}
We describe the proposed framework for a multi-robot system tracking a desired target. 
For the concepts presented, we consider Micro Aerial Vehicles (MAVs) that hover at a pre-specified height $h_{gnd}$. Furthermore, we consider 2D target destination surface. However, the proposed approaches can be extended to any 3D surface. Let there be $K$ MAVs $R_1,..., R_K$ tracking a target $x_t^P$, typically a person $P$. Each MAV computes a desired destination position $\xproj_t^{R_k}$ in the vicinity of the target position. The pose of $k^{\text{th}}$ MAV in the world frame at time $t$ is given by $\xi_t^{R_k} = [ (\x_t^{R_k})^\top ~ (\Theta_t^{R_k})^\top] \in \mathbb{R}^6$. 
Let there be $M$ obstacles in the environment $O_1,...,O_M$. The $M$ obstacles include $R_k$'s neighboring MAVs and other obstacles in the environment.

The key requirements in a multi-robot target tracking scenario are, (i) to not lose track of the moving target, and (ii) to ensure that the robots avoid other robot agents and all obstacles (static and dynamic) in their vicinity. In order to address both these objectives in an integrated approach, we formulate a formation control (FC) algorithm, as detailed in Algorithm \ref{Alg:fc}. The main steps have the following functionality, (i)  destination point computation depending on target movement, (ii) obstacle avoidance force generation, (iii)  decentralized quadratic model predictive control (DQMPC) based planner for way point generation, and (iv) a low-level position controller.

To track the waypoints generated by the MPC based planner we use a geometric tracking controller.  The controller is based on the control law proposed in~\cite{lee2010geometric},  which has a proven global convergence, aggressive maneuvering controllability and excellent position tracking performance. Here, the rotational dynamics controller is developed directly on $SO(3)$ and thereby avoids any singularities that arise in local coordinates. Since the MAVs used in this work are under-actuated systems, the desired attitude generated by the outer-loop translational dynamics is controlled by means of the inner-loop torques.

\subsection{DQMPC based Formation Planning and Control}
The goal of the formation control algorithm running on each MAV $R_k$ is to  
\begin{enumerate}
\item Hover at a pre-specified height $h_{gnd}$.
\item Maintain a distance $d^{R_k}$ to the tracked target.
\item Orient at yaw, $\psi^{R_k}$, directly facing the tracked target.
\end{enumerate}
Additionally, MAVs must adhere to the following constraints, 
\begin{enumerate}
\item To maintain a minimum distance $d_{min}$ from other MAVs as well as static and dynamic obstacles.
\item To ensure that MAVs respect the specified state limits.
\item To ensure that control inputs to MAVs are within the pre-specified saturation bounds.
\end{enumerate}
\vspace{-0.5em}
\begin{algorithm}[h]
\small
\caption{MPC-based formation controller and obstacle avoidance by MAV $R_k$ with inputs $\lbrace \x_t^{P}, ~~ \x_t^{O_j}; j = 1:M \rbrace$} 
\begin{spacing}{1.3}
\begin{algorithmic}[1]
\label{Alg:fc}
\STATE $\lbrace \xproj_t^{R_k} \rbrace \leftarrow $ Compute Destination Position $\lbrace \psi_t^{R_k}, \x_{t}^{R_k}, d^{R_k}, h_\mathrm{gnd} \rbrace$
\STATE $\left[\mathbf{f}_{t}^{R_k}(0),\dots,\mathbf{f}_{t}^{R_k}(N)\right] \leftarrow $ Obstacle Force $\lbrace {\x_{t}^{R_k}, \x_{t}^{O_j}(1:N+1),\forall j } \rbrace$
\STATE $\lbrace \x_t^{R_k*}, \dot{\x}_t^{R_k*}, \nabla J_{DQMPC} \rbrace \leftarrow $ DQMPC$\lbrace \xproj_t^{R_k}, \x_t^{R_k}, \mathbf{f}_t^{R_k}(0:N),\mathbf{g}\rbrace$
\STATE $\lbrace \psi_{t+1}^{R_k} \rbrace \leftarrow$ Compute Desired Yaw $\lbrace \x_t^{R_k}, \|\nabla J_{DQMPC} \|\rbrace$
\STATE $\mathbf{Transmit}$  $\x_t^{R_k*}(N+1),\dot{\x}_t^{R_k*}(N+1),  \psi_{t+1}^{R_k}$ to Low-level Controller
\normalsize
\end{algorithmic}
\end{spacing}
\end{algorithm} 
\vspace{-1em}
Algorithm \ref{Alg:fc} outlines the strategy used by each MAV $R_k$ at every  discrete time instant $t$. In line 1, MAV $R_k$ computes its desired position $\xproj_t^{R_k}$ on the desired surface using simple trigonometry. For example, if the desired surface is a circle, centered around the target location $\x_{t}^{P}$ with a radius $d^{R_k}=constant \ \forall R_k$, then the desired position for time instant $t$ is given by $\xproj_t^{R_k} = \x_{t}^{P}+ \left[d^{R_k}cos(\psi_t^{R_k}) ~~ d^{R_k}sin(\psi_t^{R_k}) ~~ h_{gnd}\right]^{\top}$. Here $\psi_t^{R_k}$ is the yaw of $R_k$ w.r.t. the target. It is important to note that the distance $d^{R_k}$ is an input to the DQMPC and is not necessarily the same for each MAV. 

In line 2, an input potential field force vector $\left[\mathbf{f}_{t}^{R_k}(0),\dots,\mathbf{f}_{t}^{R_k}(N)\right] ^{\top} \in \mathbb{R}^{3 \times (N+1)}$ is computed for a planning horizon of $(N+1)$ discrete time steps by using the shared trajectories from other MAVs and positions of obstacles in the vicinity. If no trajectory information is available, the instantaneous position based potential field force value is used for the entire horizon. Section \ref{potential_field} details the numerical computation of these field force vectors.

In line 3, an MPC based planner solves a convex optimization problem (DQMPC) for a planning horizon of $(N+1)$ discrete time steps. We consider nominal accelerations $[\ao_t^{R_k}(0) \cdots \ao_t^{R_k}(N)]^\top \in \mathbb{R}^{3 \times (N+1)}$ as control inputs to DQMPC. The accelerations describe the 3D translational motion of $R_k$.
\begin{equation}
 \ao_t^{R_k}(n) = ~ \ddot{\x}_t^{R_k}(n)
\end{equation}
where $n$ is the current horizon step. 
\setlength{\belowcaptionskip}{0pt}
\begin{figure*}[t]
	\centering
	\begin{subfigure}[t]{0.24\textwidth}        
		\includegraphics[scale=0.4]{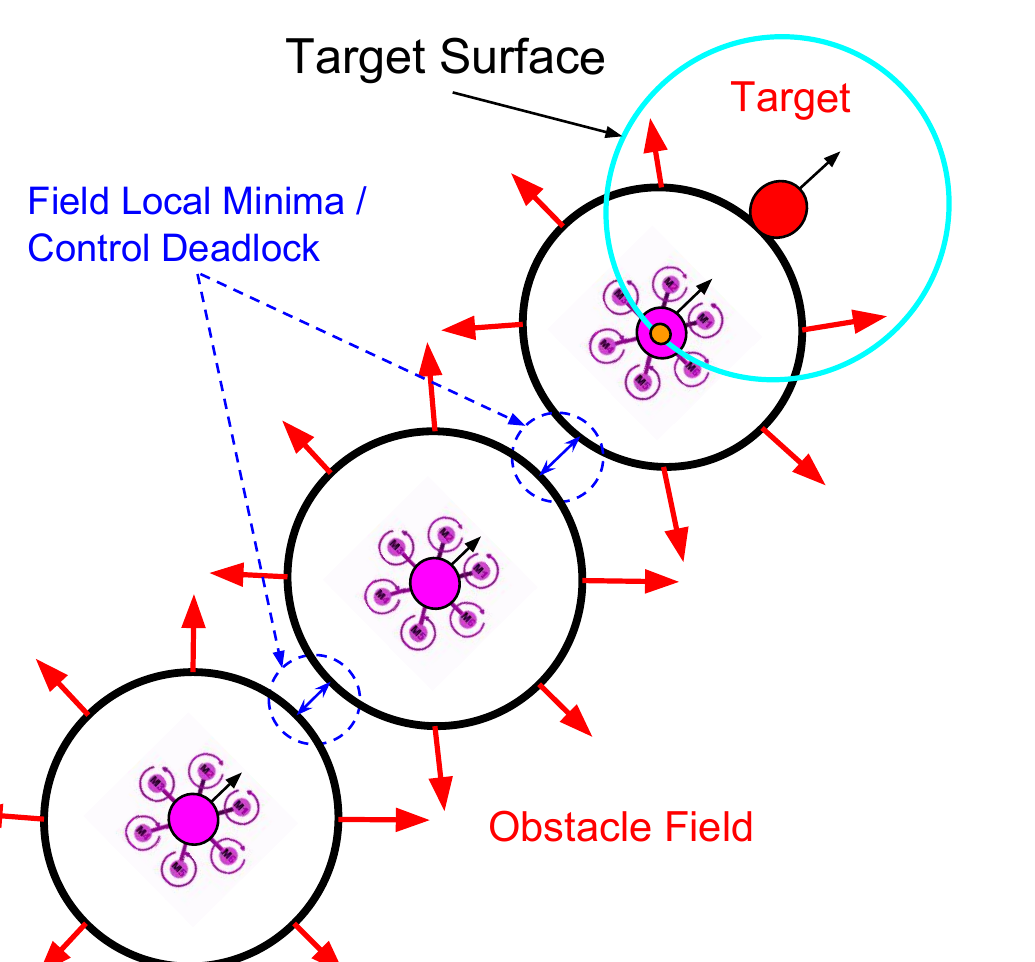}   
		\caption{\scriptsize{Field local minima problem}}
	\end{subfigure}
	\begin{subfigure}[t]{0.24\textwidth}        
		\includegraphics[scale=0.4]{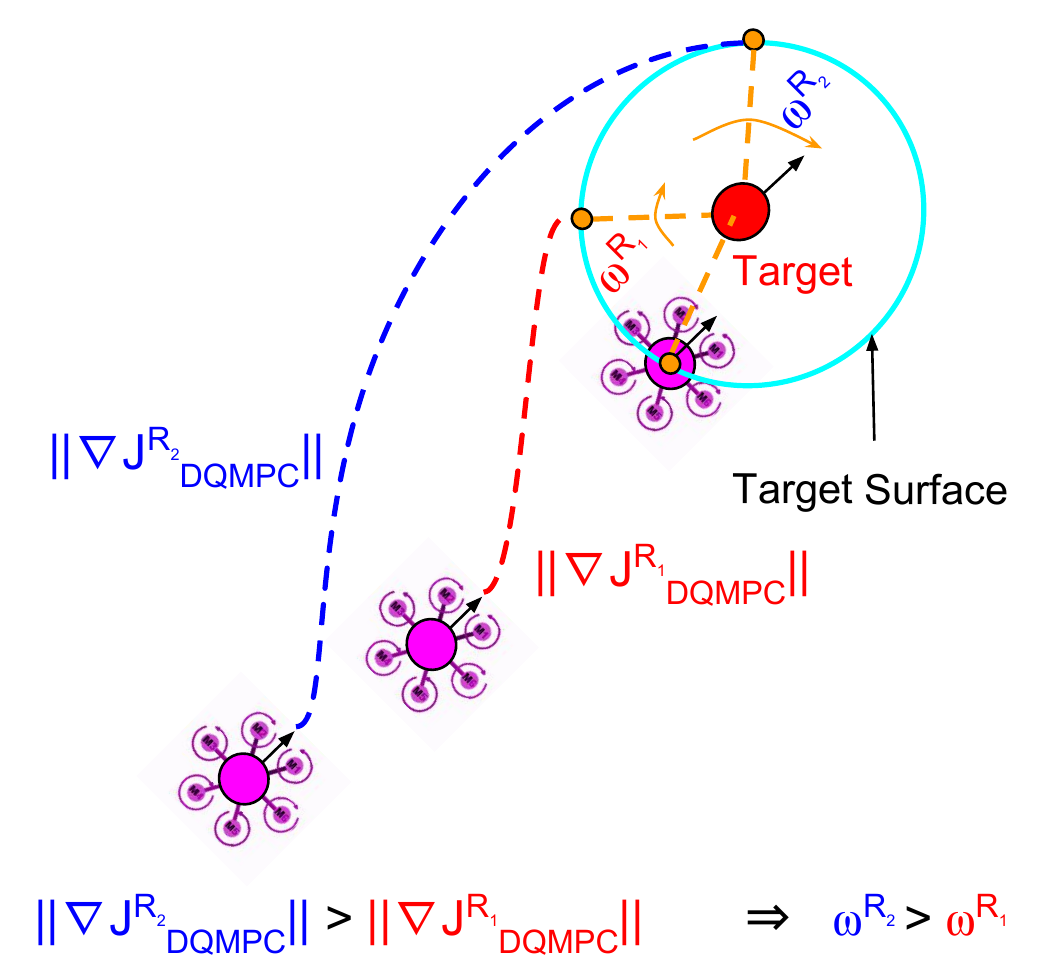}
		\caption{\scriptsize{Swivelling Robot Destination Method.}}
	\end{subfigure} 
	\begin{subfigure}[t]{0.24\textwidth}
		\centering
		\includegraphics[scale=0.4]{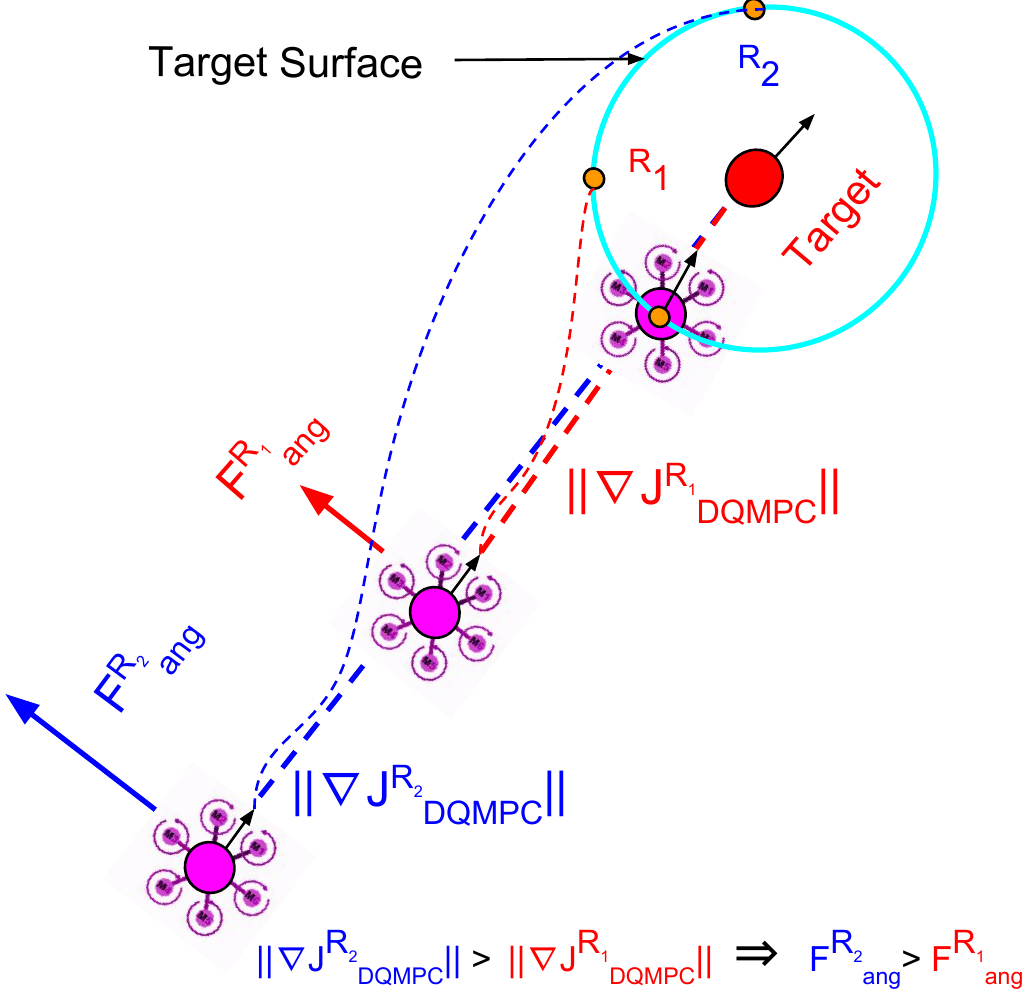}  
		\caption{\scriptsize{Approach Angle Method}}
	\end{subfigure}
	\begin{subfigure}[t]{0.24\textwidth}
		\centering
		\includegraphics[scale=0.4]{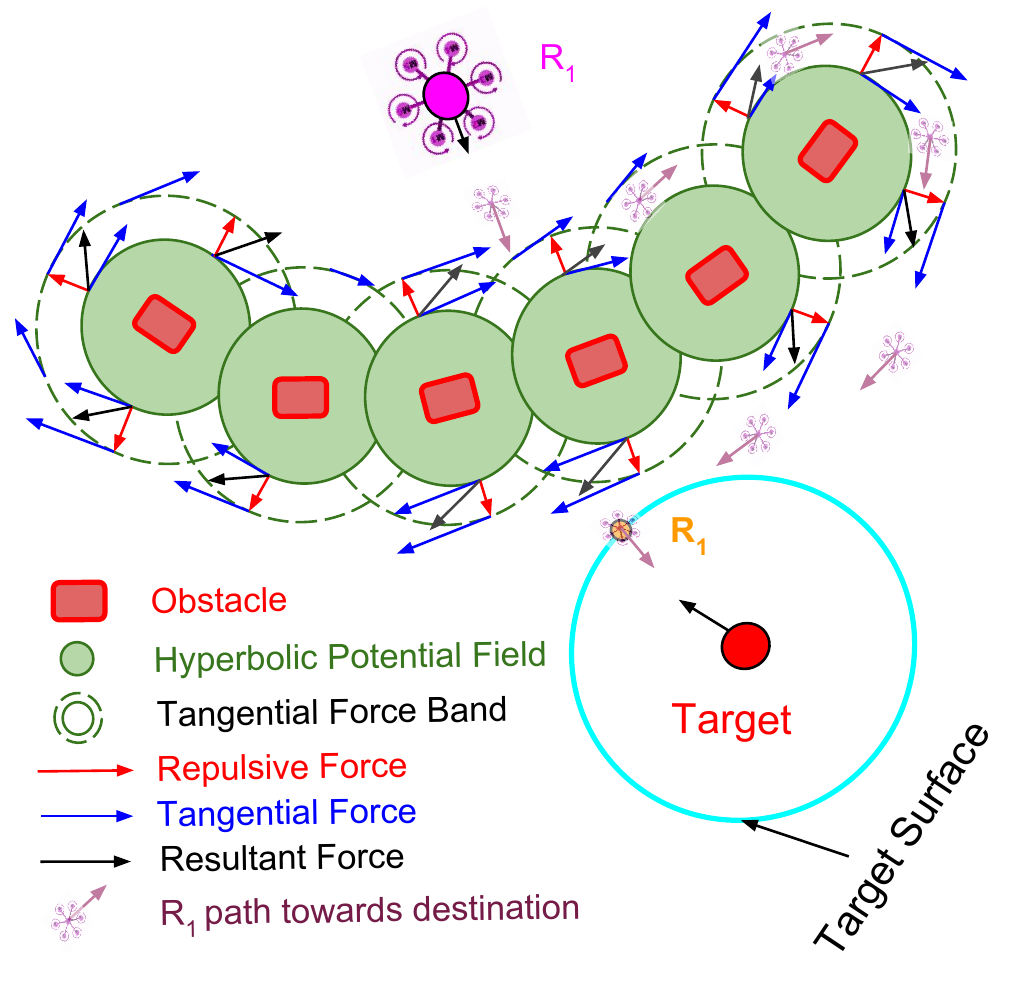}
		\caption{\scriptsize{Tangential Band Method}}
	\end{subfigure}
	\caption{Illustration of the field local minima problem in obstacle avoidance and different proposed methods.} 
	\label{proposedApproachfig}
	\vspace{-1em}
\end{figure*}
\setlength{\belowcaptionskip}{-5pt}
The state vector of the discrete-time DQMPC consists of $R_k$'s position $\x_t^{R_k}(n) \in \mathbb{R}^3$ and  velocity $\dot{\x}_t^{R_k}(n)\in \mathbb{R}^3$. 
The optimization objective is,
\footnotesize
\begin{align} \label{J_DQMPC}
J_{\mathrm{DQMPC}} =&  \Big(\sum_{n=0}^{N} \big( {\bm{\Omega_{i}}} (\ao_t^{R_k}(n)+\mathbf{f}_{t}^{R_k}(n)+\bm{g})^2\big) + \nonumber \\
& \bm{\Omega_{t}} \big(\left[\x_t^{R_k}(N+1)^{\top} ~~ \dot{\x}_t^{R_k}(N+1)^{\top} \right] - \left[ (\xproj_t^{R_k})^{\top}  ~~ \mathbf{0}^{\top}\right]\big)^2 \Big)
\end{align}
\normalsize
The optimization is defined by the following equations. \small
\begin{equation} \label{DQMPC}
\x(1)_t^{R_k*}\dots\x(N+1)_t^{R_k*},\ao_t^{R_k*}(0)\dots\ao_t^{R_k*}(N) = \argmin{\ao_t^{R_k}(0)\dots\ao_t^{R_k}(N)} (J_{DQMPC})
\end{equation} 
subject to, 
\begin{align}
& \hspace{-0.7cm}\left[\x_t^{R_k}(n+1)^{\top} ~ \dot{\x}_t^{R_k}(n+1)^{\top}\right]^\top =  \label{state-space}\nonumber\\
& \quad\quad\quad\mathbf{A}\left[\x_t^{R_k}(n)^{\top} ~ \dot{\x}_t^{R_k}(n)^{\top}\right]^\top + \mathbf{B}(\ao_t^{R_k}(n)+\mathbf{f}_{t}^{R_k}(n)+\bm{g}), \\
& \ao_\mathrm{min} \leq ~ \ao_t^{R_k}(n) \leq \ao_\mathrm{max}, \\
& \x_\mathrm{min} \leq ~ \x_t^{R_k}(n) \leq ~ \x_\mathrm{max}, \\
& \dot{\x}_\mathrm{min} \leq ~ \dot{\x}_t^{R_k}(n) \leq ~ \dot{\x}_\mathrm{max} \label{last_DQMPC}
 \end{align}
 \normalsize
where, $\bm{\Omega_{i}}$ and $\bm{\Omega_{t}}$ are positive definite weight matrices for input cost and terminal state (computed desired position $\xproj_t^{R_k}$ and desired velocity $\dot{\xproj}_t^{R_k} = 0$) respectively, 
$\mathbf{f}_{t}^{R_k}(n)$ is the pre-computed external obstacle force, $\bm{g}$ is the constant gravity vector. The discrete-time state-space evolution of the robot is given by \eqref{state-space}. The  dynamics ($\mathbf{A} \in \mathbb{R}^{3\times3}$) and control transfer ($\mathbf{B} \in \mathbb{R}^{3\times3}$) matrices are given by,
\begin{equation}
\mathbf{A}=
  \begin{bmatrix}
    \mathbf{I}_3 & \Delta t\mathbf{I}_3 \\
    \mathbf{0}_3 & \mathbf{I}_3
  \end{bmatrix}, ~
\mathbf{B}=
  \begin{bmatrix}
    \frac{\Delta t^2}{2}\mathbf{I}_3 \\
    \Delta t\mathbf{I}_3
  \end{bmatrix}.
\end{equation}
where, $\Delta t$ is the sampling time. The quadratic program generates optimal control inputs $\left[\ao_t^{R_k}(0) \cdots \ao_t^{R_k}(N) \right]$ and the corresponding trajectory $\left[ \x_t^{R_k}(1) ~ \dot{\x}_t^{R_k}(1)  \cdots \x_t^{R_k}(N+1) ~ \dot{\x}_t^{R_k}(N+1) \right]$ towards the desired position. The final predicted position and velocity of the horizon $\x_t^{R_k*}(N+1),\dot{\x}_t^{R_k*}(N+1)$ is used as  desired input to the low-level flight controller. 
The MPC based planner avoids obstacles (static and dynamic) through pre-computed horizon potential force $\mathbf{f}_{t}^{R_k}(n)$. This force is applied as an external control input component to the state-space evolution equation, thereby, preserving the optimization convexity. Previous methods in literature consider non-linear potential functions within the MPC formulation, thereby, making optimization non-convex and computationally expensive. 

In the next step (line 4 of Algorithm \ref{Alg:fc}), the desired yaw is computed as $\psi_{t+1}^{R_k} = atan2\big(\frac{y_t^P - y_t^{R_k}}{x_t^P - x_t^{R_k}}\big)$. This  describes the angle with respect to the target position $\x_t^P$ from the MAV's current position $\x_t^{R_k}$. 
The way-point commands consisting of the position and desired yaw angle are sent to the low-level flight  position controller.
Although no specific robot formation geometry is enforced, the DQMPC naturally results in a dynamic formation depending on desired destination surface.

\subsection{Handling Non-Convex Collision Avoidance Constraints} \label{potential_field}
In our approach, at any given point there are two forces acting on each robot, namely (i) the attractive force due to the optimization objective (eq.\eqref{DQMPC}), and (ii) the repulsive force due to the potential field around obstacles ($\mathbf{f}_{t}^{R_k}(n)$). In general, a repulsive force can be modeled as a force vector based on the distance w.r.t. obstacles. Here, we have considered the potential field force variation as a hyperbolic function ($F_{hyp}^{R_k,O_j}(d(n))$) of distance between MAV $R_k$ and obstacle $O_j$. We use the formulation in \cite{secchi2013bilateral} to model $F_{hyp}^{R_k,O_j}(d(n))$. Here, $d(n) = \|\x_{t-1}^{R_k}(n) - \x_t^{O_j}(n)\|_2, ~ \forall n \in [0,\dots,N]$ is the distance between the MAV's predicited horizon positions from the previous time step $(t-1)$ and the obstacles (which includes shared horizon predictions of other MAVs).  The repulsive force vector is,
\begin{equation}\label{repulsive}
\bm{F}_{rep}^{R_k,O_j}(n) = 
\begin{cases}
F_{hyp}^{R_k,O_j}(d(n))\;\mathbf{\alpha},& \text{if} ~~ d(n) < d_{safe} \\
0, & \text{if} ~~ d(n) > d_{safe}
\end{cases}\;,
\end{equation}
where, $d_{safe}$ is the distance from the obstacle where the potential field magnitude is non-zero. 
 $\mathbf{\alpha} =  \frac{\x_t^{O_j}(n)-\x_{t-1}^{R_k}(n)}{\|\x_t^{O_j}(n)-\x_{t-1}^{R_k}(n)\|_2}$ is the unit vector in the direction away from the obstacle.  Additionally we consider a distance $d_{min} << d_{safe}$ around the obstacle, where the potential field magnitude tends to infinity. 
The potential force acting on an agent per horizon step $n$ is,
\begin{equation} \label{totalForce}
\mathbf{f}_t^{R_k}(n)= \sum_{\forall \ j }^{}\mathbf{F}_{rep}^{R_k,O_j}(n),
\end{equation}
which is added into the system dynamics in eq.\eqref{state-space}.
\vspace{-0.5em}

\subsection{Resolving the Field Local Minima Problem}
The key challenge in potential field based approaches is the field local minima issue \cite{koren1991potential}. When the summation of attractive and repulsive forces acting on the robot is a zero vector, the robot encounters field local minima problem \footnote{note that this is different from optimization objective's local minima.}. Equivalently, a control deadlock could also arise when the robot is constantly pushed in the exact opposite direction.
Both local minima and control deadlock are undesirable scenarios.
From  equation \eqref{state-space} and Algorithm \ref{Alg:fc}, it is clear that the  optimization can characterize control inputs that will not lead to collisions, but, cannot characterize those control inputs that lead to these scenarios.
In such cases the gradient of optimization would be non-zero indicating that the robot knows its direction of motion towards the target, but cannot reach the destination surface because the potenial field functions are not directly used in DQMPC constraints.
Here, we propose three methodologies for field local minima avoidance.

\subsubsection{Swivelling Robot Destination (SRD) method} \label{diminishingOmega}
This method is based on the idea that the MAV destination $\xproj_t^{R_k}$ is an external input to the optimization. Therefore, each MAV can change its $\xproj_t^{R_k}$ to push itself out of field local minima. For example, consider the scenario shown in Fig. \ref{proposedApproachfig}(a), where three robots are axially aligned towards the target. Since the angles of approach are equal, the desired destination positions are the same for $R_1$ and $R_2$, i.e., $\xproj_t^{R_1} = \xproj_t^{R_2}$. This results in temporary deadlock and will slow the convergence to desired surface. We construct the SRD method to solve this deadlock problem as follows: (i) the gradient of DQMPC objective of $R_k$ is computed, (ii)  a swivelling velocity $\omega^{R_k}$ is calculated based on the magnitude of gradient, and (iii) $\xproj_t^{R_k}$ swivels by a distance proportional to $\omega^{R_k}$ as shown in Fig.\ref{proposedApproachfig}(b). This ensures that the velocities at which each $\xproj_t^{R_k}$ swivels is different until the robot reaches the target surface, where the gradient tends to zero. The gradient of the optimization with respect to the last horizon step control and state vectors, is computed as follows.

\small
\begin{equation}
\frac{\partial J_{DQMPC}}{\x_t^{R_k}(N+1)} = 2 \bm{\Omega_{t}} (\left[\x_t^{R_k}(N+1)^{\top} ~~ (\dot{\x}_t^{R_k}(N+1))^{\top} \right] - \left[ (\xproj_t^{R_k})^{\top}  ~~ \mathbf{0}^{\top}\right])^{\top} \nonumber
\end{equation}
\begin{equation}
\frac{\partial J_{DQMPC}}{\ao_t^{R_k}(N)} = 2 \bm{\Omega_{i}}(\ao_t^{R_k}(n)+\mathbf{f}_{t}^{R_k}(n)+\bm{g}) + 
2 \bm{\Omega_{t}}B(\x_t^{R_k}(N+1)-\xproj_t^{R_k}) \nonumber
\end{equation}
\begin{equation}
\nabla J_{DQMPC}^{R_k} = \frac{\partial J_{DQMPC}}{\x_t^{R_k}(N+1)} + \frac{\partial J_{DQMPC}}{\ao_t^{R_k}(N)}.
\end{equation}
\normalsize

For circular target surface, the destination point swivel rate is,
\small
\begin{eqnarray}
\xproj_t^{R_k} =& \x_{t}^{P}+ \begin{bmatrix} d^{R_k}cos(\psi_t^{R_k} \pm k_s \|\nabla J_{DQMPC}^{R_k}\|) \\ d^{R_k}sin(\psi_t^{R_k}\pm k_s\|\nabla J_{DQMPC}^{R_k}\|) \\ h_{gnd} \end{bmatrix} ^\top 
\end{eqnarray}
\normalsize
where, $k_s$ is a user-defined gain controlling the impact of $\|\nabla J_{DQMPC}^{R_k}\|$. The swivel direction of each ${R_k}$ is decided by its approach direction to target. Positive and negative $\psi_t^{R_k}$ leads to a clockwise and anti-clockwise swivel respectively. 

\subsubsection{Approach Angle Towards Target Method} \label{angle_of_approach}
In this method, the local minima and control deadlock is addressed by including an additional potential field function which depends on the approach angle of the robots towards the target.
Here, we (i) compute the approach angle of robot $R_k$ w.r.t. the target, (ii) compute the gradient of the objective, and (iii) compute a force $F_{ang}^{R_k,O_j}$ in the direction normal to the angle of approach, as shown in Fig. \ref{proposedApproachfig}(c). The magnitude of  $F_{ang}^{R_k,O_j}$ depends on the sum of gradients $\nabla J_{DQMPC}^{R_k}$ and the hyperbolic function (see Sec. \ref{potential_field}) between the approach angles of robot $R_k$ and obstacles $O_j$ w.r.t. the target. This potential field force is computed as,
\begin{equation}
\mathbf{F}_{ang}^{R_k,O_j}(n) = \nabla J_{DQMPC}^{R_k} \;F_{hyp}^{R_k,O_j}((\theta^{R_k}(n)-\theta^{O_j}(n))^2)\mathbf{\hat{\beta}}  \;\;\forall j 
\end{equation}
\begin{equation}
\mathbf{\beta} = \pm \frac{\x_t^{R_k}(n)-\xproj_t^{R_k}}{\|\x_t^{R_k}(n)-\xproj_t^{R_k}\|_2} \; ;\quad \mathbf{\hat{\beta}}.\mathbf{\beta} = 0.
\end{equation}
Here $\theta^{R_k}(n)$ and $\theta^{O_j}(n)$ are the angles of $R_k$ and obstacle $O_j$ with respect to the target. The angles $\theta^{O_j} \ \forall \ j$ w.r.t target are computed by each $R_k$, as part of the force pre-computation using $O_j$'s position. $\beta$ and $\hat{\beta}$ are the unit vectors in the approach direction to the target and its orthogonal respectively, with $\pm$ dependent on $\theta^{R_k}$ w.r.t. the target. 
The $\mathbf{f}_{t}^{R_k}(n)$ for $n^{th}$ horizon step is therefore
\begin{equation}
\mathbf{f}_{t}^{R_k}(n) = \sum_{\forall \ j}\mathbf{F}_{rep}^{R_k,O_j}(n)+\mathbf{F}_{ang}^{R_k,O_j}(n).
\end{equation}
Notice that the non-linear constraint of the two approach angles not being equal is converted into an equivalent convex constraint using pre-computed force values. This method ensures collision avoidance in the presence of obstacles and fast convergence to the desired target, because the net potential force direction is always away from the obstacle. 


\subsubsection{Tangential Band Method} \label{tangential_field}
The previous methods at times, do not facilitate target surface convergence because of field local minima and control deadlock. For example, static obstacles forming a U-shaped boundary between the target surface and $R_k$'s position, as shown in Fig.\ref{proposedApproachfig}(d). If the target surface is smaller than the projection of the static obstacle along the direction of approach to the desired surface, the planned trajectory is occluded. Therefore, the SRD method cannot find a feasible direction for motion. Furthermore,  angle of approach field acts only when the $\theta^{R_k}$ and $\theta^{O_j}$ are equal w.r.t. the target.

In order to resolve this, we construct a band around each obstacle where an instantaneous (at $n=0$) tangential force acts about the obstacle center. The width of this band is $ > (\dot{\x}_{max}^ {\top}\Delta t + u_{max} \frac{\Delta t^2}{2}) $ and therefore, the robot cannot tunnel out of this band within one time step $\Delta t$. This makes sure that once the robot enters  tangential band, it exits only after it has overcome the static obstacles. The direction depends on $R_k$'s approach towards the target, resulting in clockwise or anti-clockwise force based on $-ve$ or $+ve$ value of $\psi_t^{R_k}$ respectively. The outer surface of the band has only the tangential force effect, while the inner surface has both tangential and repulsive force (repulsive hyperbolic field) effects on $R_k$.
 
Within the band the diagonal entries of the positive definite weight matrix $\bm{\Omega}_{t}$ are reduced to a very low value ($\prec \prec \Omega_{t,max}$).  This ensures that the attraction field on the robot is reduced while it is being pushed away from the obstacle. Consequently, the effect of tangential force is higher in the presence of obstacles. Once the robot is out of the tangential field band (i.e., clears the U-shaped obstacles), the high weight of the $\bm{\Omega}_{t}$ is restored and the robot converges to its desired destination. Fig. \ref{proposedApproachfig}(d) illustrates this method.
The tangential force is,
\begin{equation}
\mathbf{F}_{tang}^{R_k,O_j}(0) = k_{tang}\nabla J_{DQMPC}^{R_k}\; \mathbf{\hat{\alpha}}, 
\end{equation}
where $k_{tang}$ is user-defined gain and $\mathbf{\hat{\alpha}}$ is defined s.t. $\mathbf{\pm\;\hat{\alpha}.\alpha} = 0$. The weight matrix and step horizon potential are therefore,
\begin{equation}
\bm{\Omega_{t}} = \bm{\Omega_{t,min}}, ~~ \text{if } \x_t^{R_k} \leq d(0)+d_{band} 
\end{equation}
\begin{equation}
\mathbf{f}_{t}^{R_k}(n) = \sum_{\forall j} \mathbf{F}_{rep}^{R_k,O_j}(n) + \mathbf{F}_{tang}^{R_k,O_j}(0)\;,
\end{equation}
where $d_{band}$ is the tangential band width. The values of the weights can vary between $\bm{\Omega_{t,min}} \prec \bm{\Omega_{t}} \prec \bm{\Omega_{t,max}}$ and changes only when the robot is within the influence of tangential field of any obstacle.
In summary, the tangential band method not only guarantees collision avoidance for any obstacles but also facilitates robot convergence to the target surface. 
In rare scenarios, e.g., when the static obstacle almost encircles the robots and if the desired target surface is beyond such an obstacle, the robots could get trapped in a loop within the tangential band. This is because a minimum attraction field  towards the target always exists. Since in this work,  the objective is local motion planning in dynamic environments with no global information and map, we do not plan for a feasible trajectory out of such situations. 

\begin{figure}
	\centering
	    \begin{subfigure}[t]{0.24\textwidth}        
	    	\includegraphics[scale=0.27]{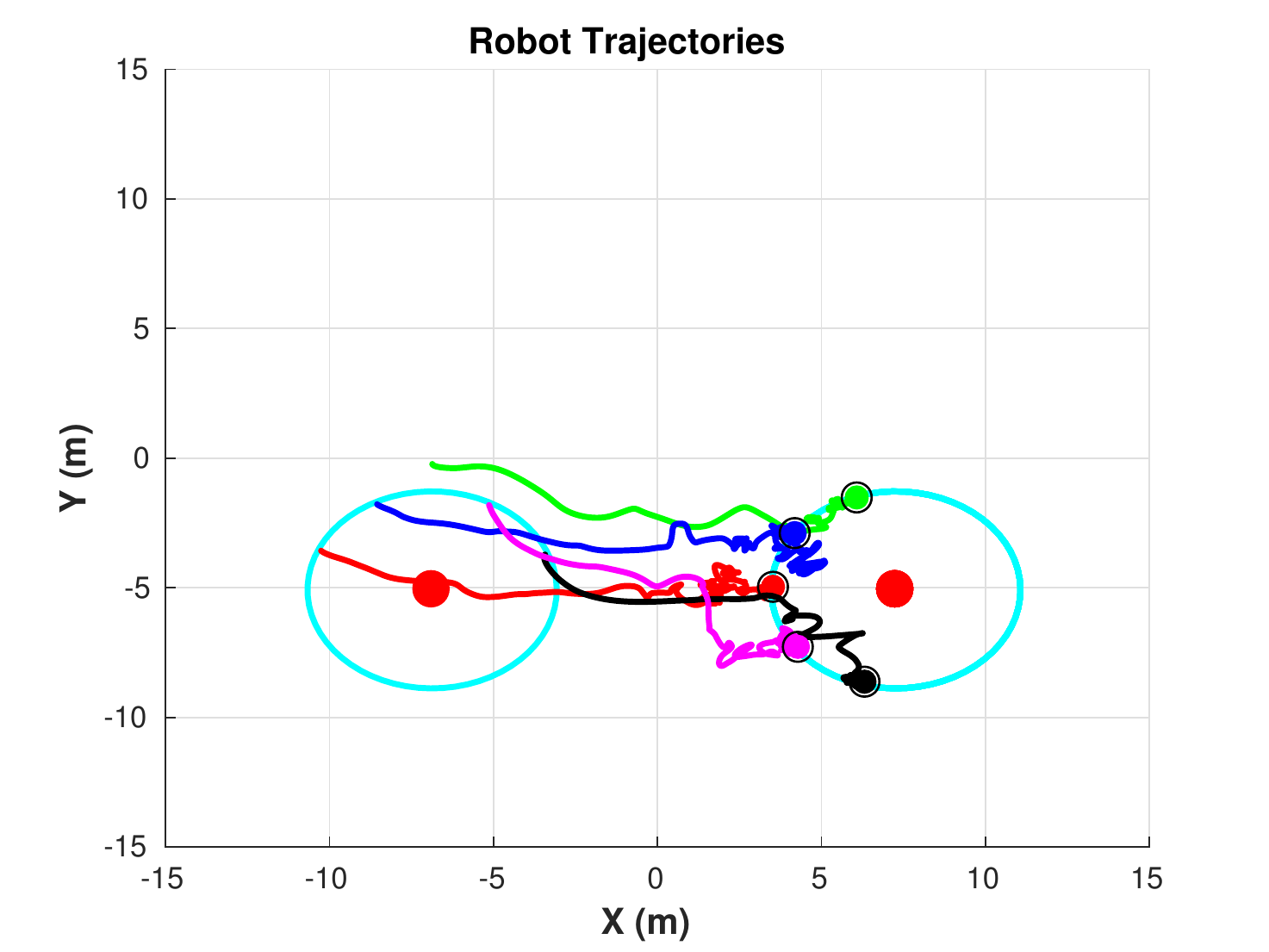}   
	    	\caption{}     
	    	\label{DQMPC_5}
	    \end{subfigure}
	    \begin{subfigure}[t]{0.24\textwidth}        
	    	\includegraphics[scale=0.22]{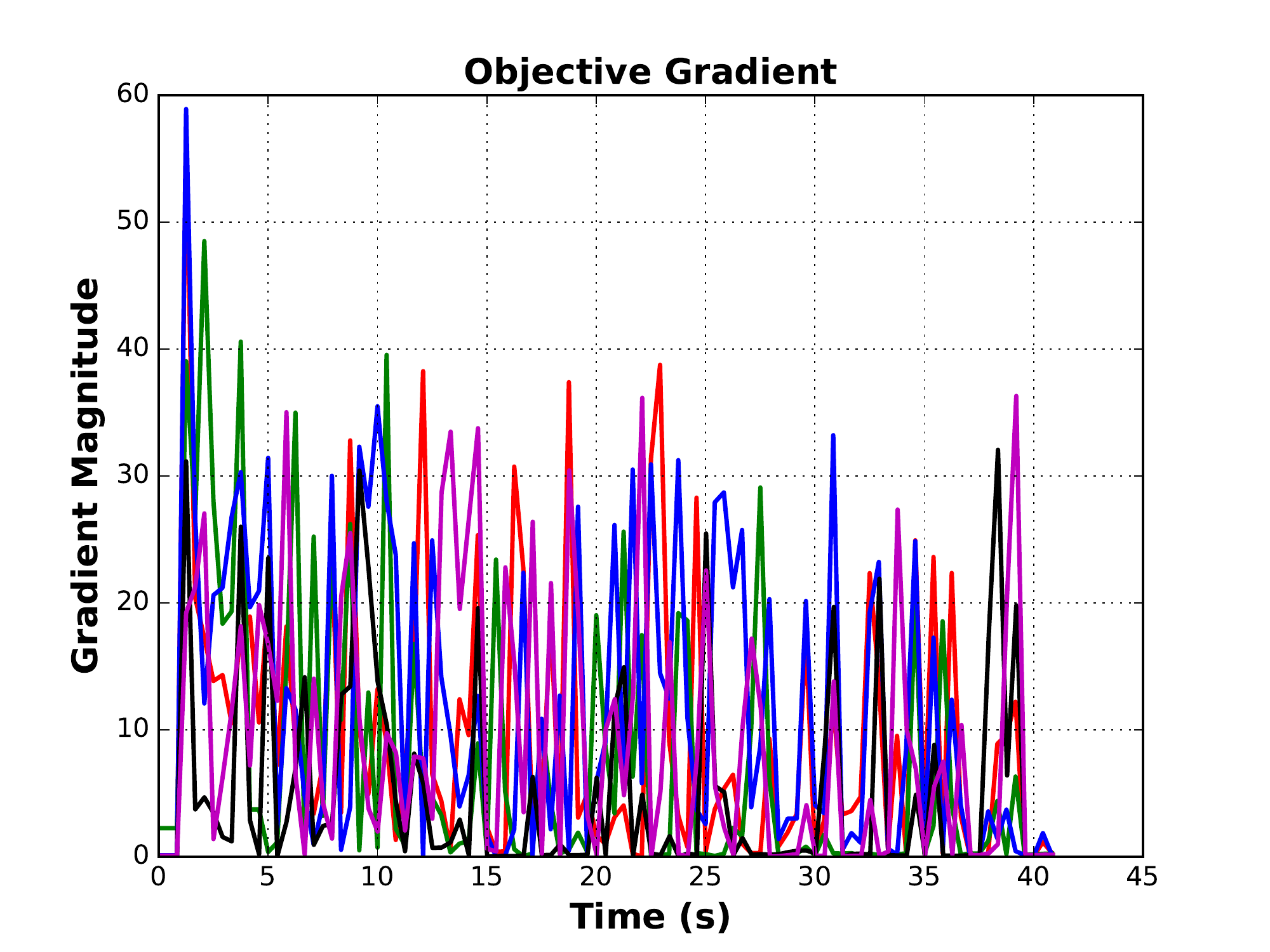}
	    	\caption{}
	    	\vspace{0.1cm}
	    	\label{DQMPC_5_grad}
	    \end{subfigure} 
	    \begin{subfigure}[t]{0.24\textwidth}
	    	\centering
	    	\includegraphics[scale=0.27]{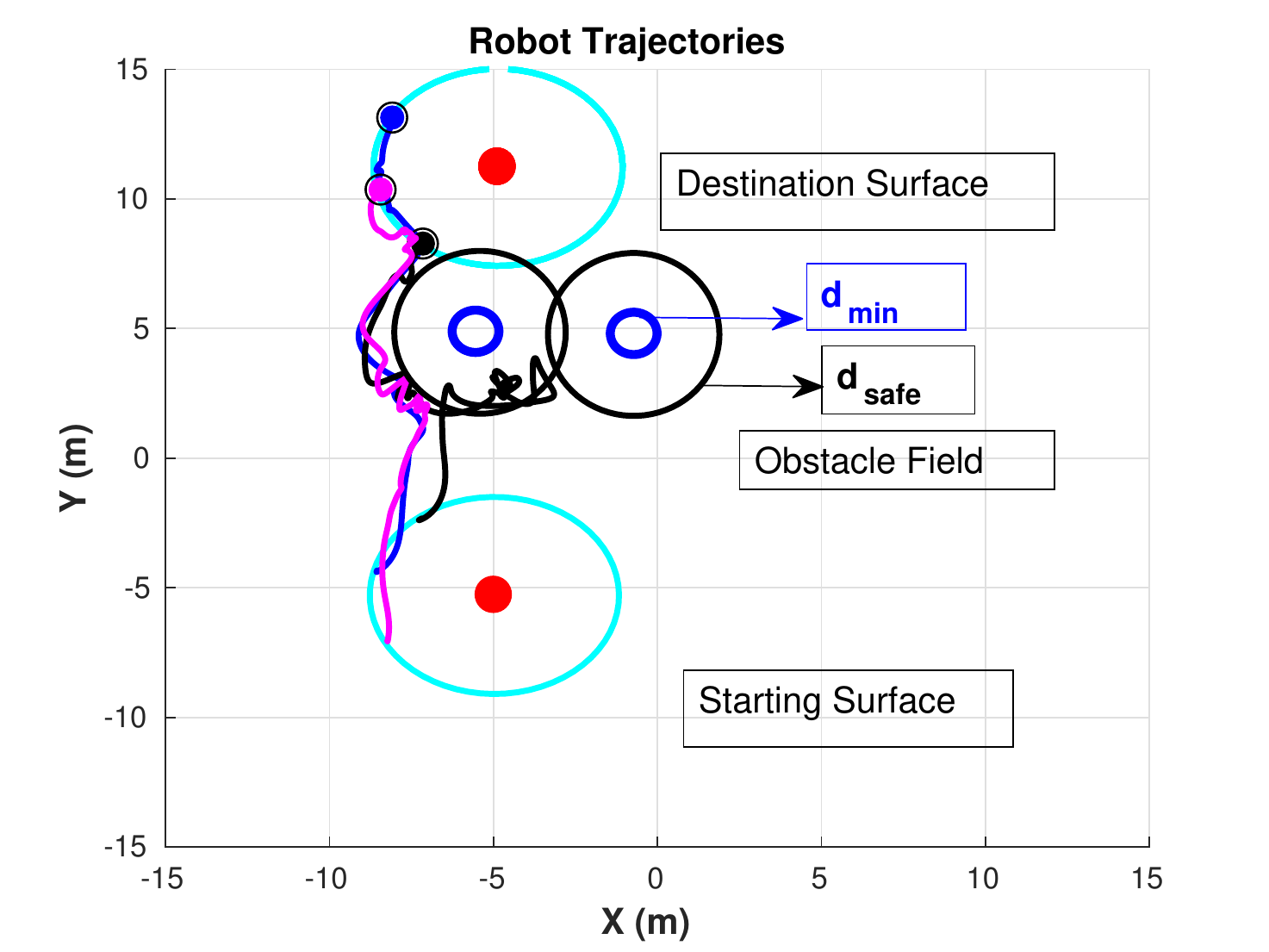}  
	    	\caption{}    
	    	\label{DQMPC_32}
	    \end{subfigure}
	    \begin{subfigure}[t]{0.24\textwidth}
	    	\centering
	    	\includegraphics[scale=0.21]{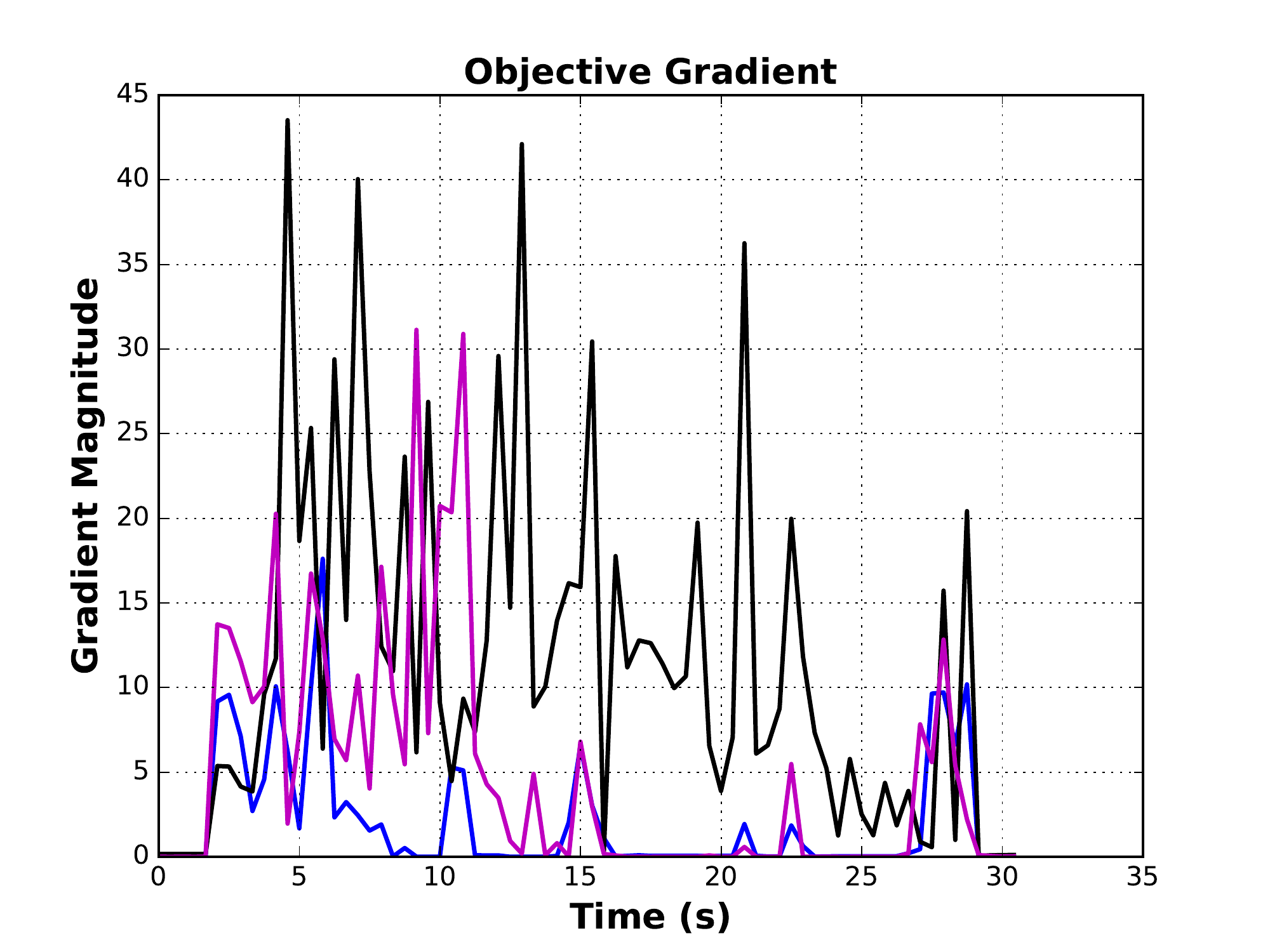}
	    	\caption{}
	    	\vspace{0.1cm}
	    	\label{DQMPC_32_grad}
	    \end{subfigure}
		\begin{subfigure}[t]{0.24\textwidth}
			\centering
			\includegraphics[scale=0.27]{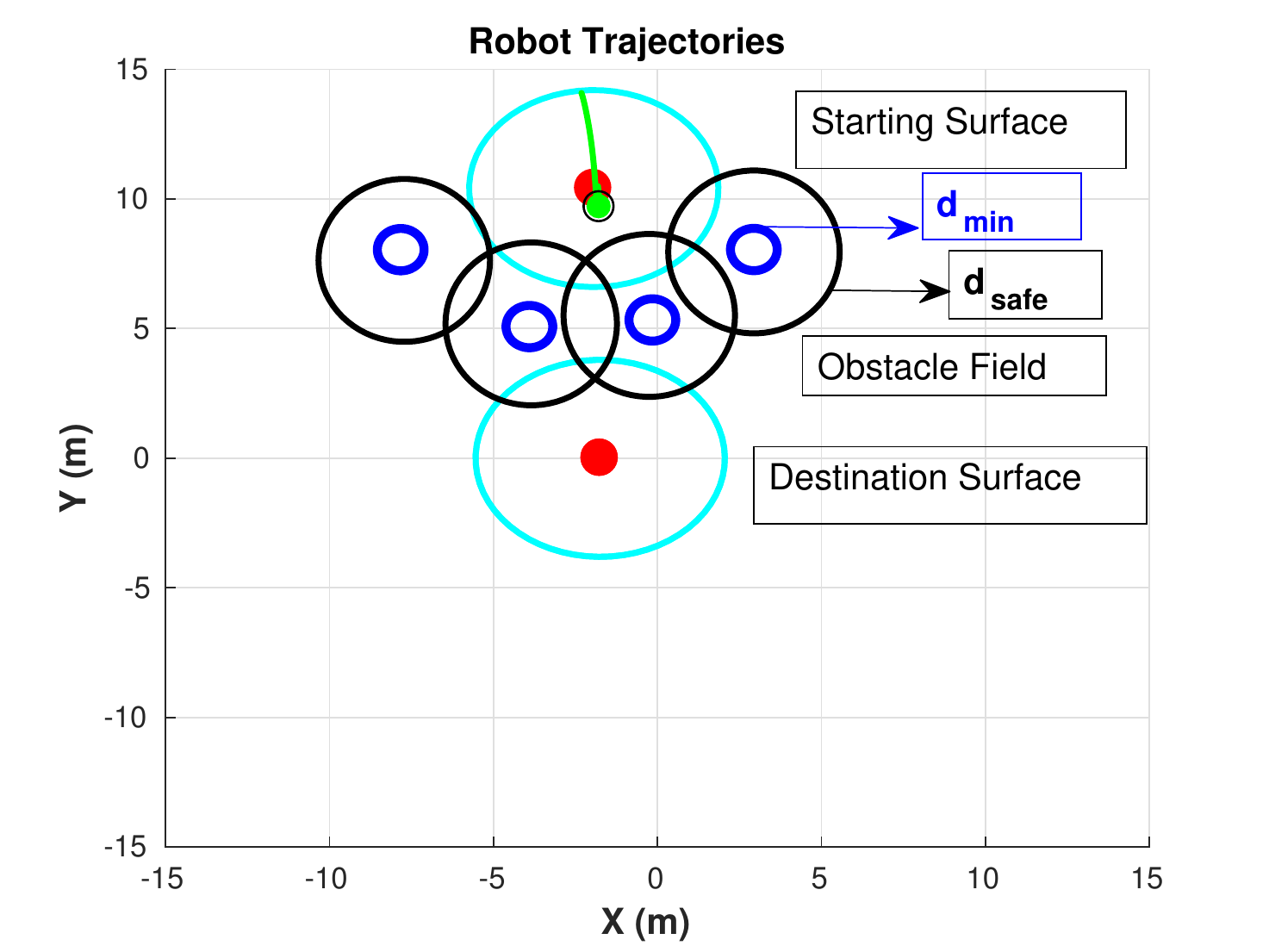} 
			\caption{}   	
			\label{DQMPC_14}
		\end{subfigure}
		\begin{subfigure}[t]{0.24\textwidth}
			\centering
			\includegraphics[scale=0.22]{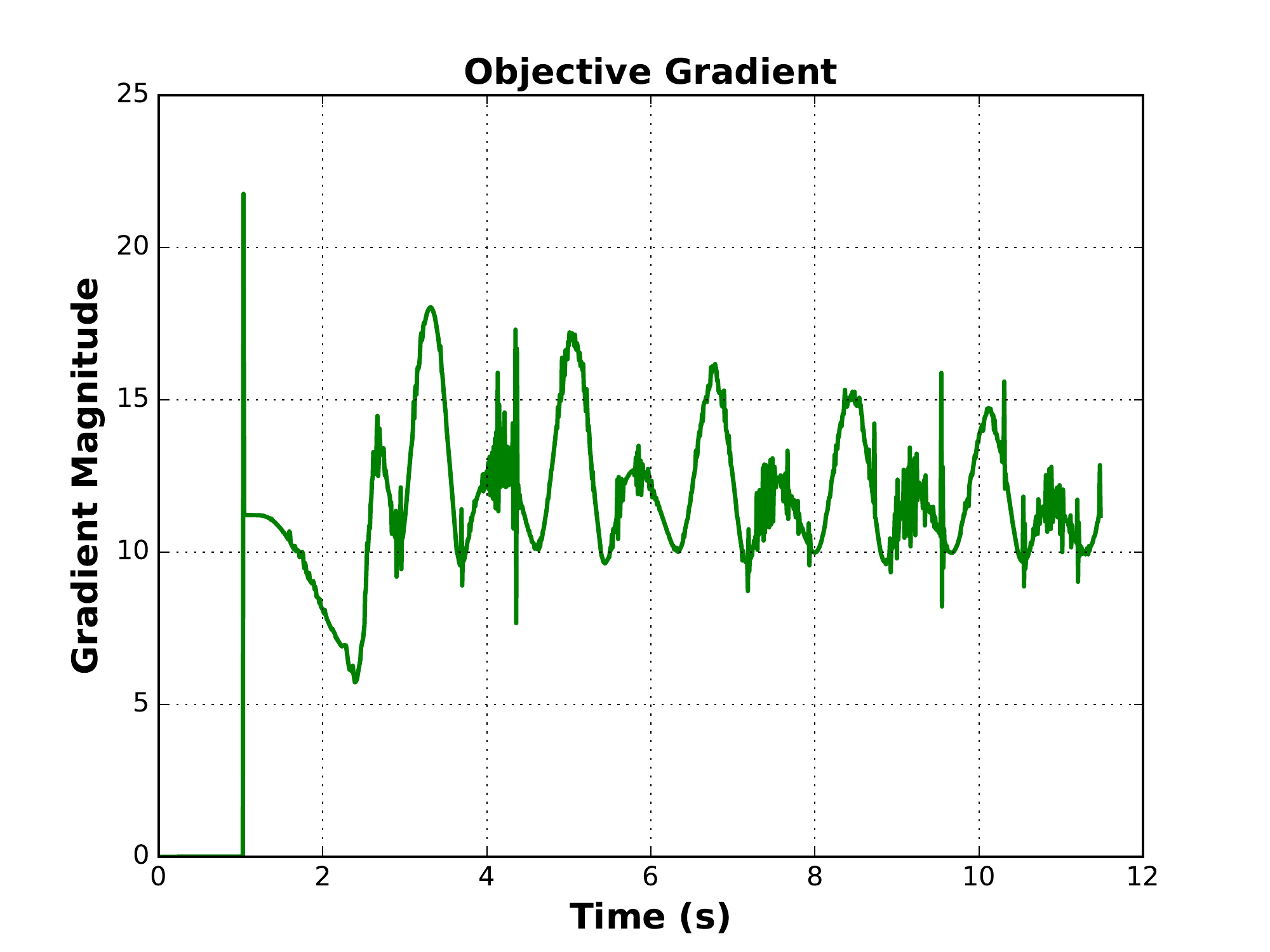}
			\caption{}
			\label{DQMPC_14_grad}
		\end{subfigure}      
		\caption{MAV trajectories and optimization gradients of baseline DQMPC optimization. The colors (red, green, blue, black, magenta) represent $R_k$s and their respective gradients.}
		\label{DQMPC_Result} 
		\vspace{-1em}
\end{figure}

\section{Results and Discussions}
\label{sec:results} 
In this section, we detail the experimental setup and the results of our DQMPC based approach along with  the field local minima resolving methods proposed in Sec.~\ref{sec:proposedappraoch} for obstacle avoidance and reaching the target surface.
\begin{figure}
	\centering
	\begin{subfigure}[t]{0.24\textwidth}        
		\includegraphics[scale=0.27]{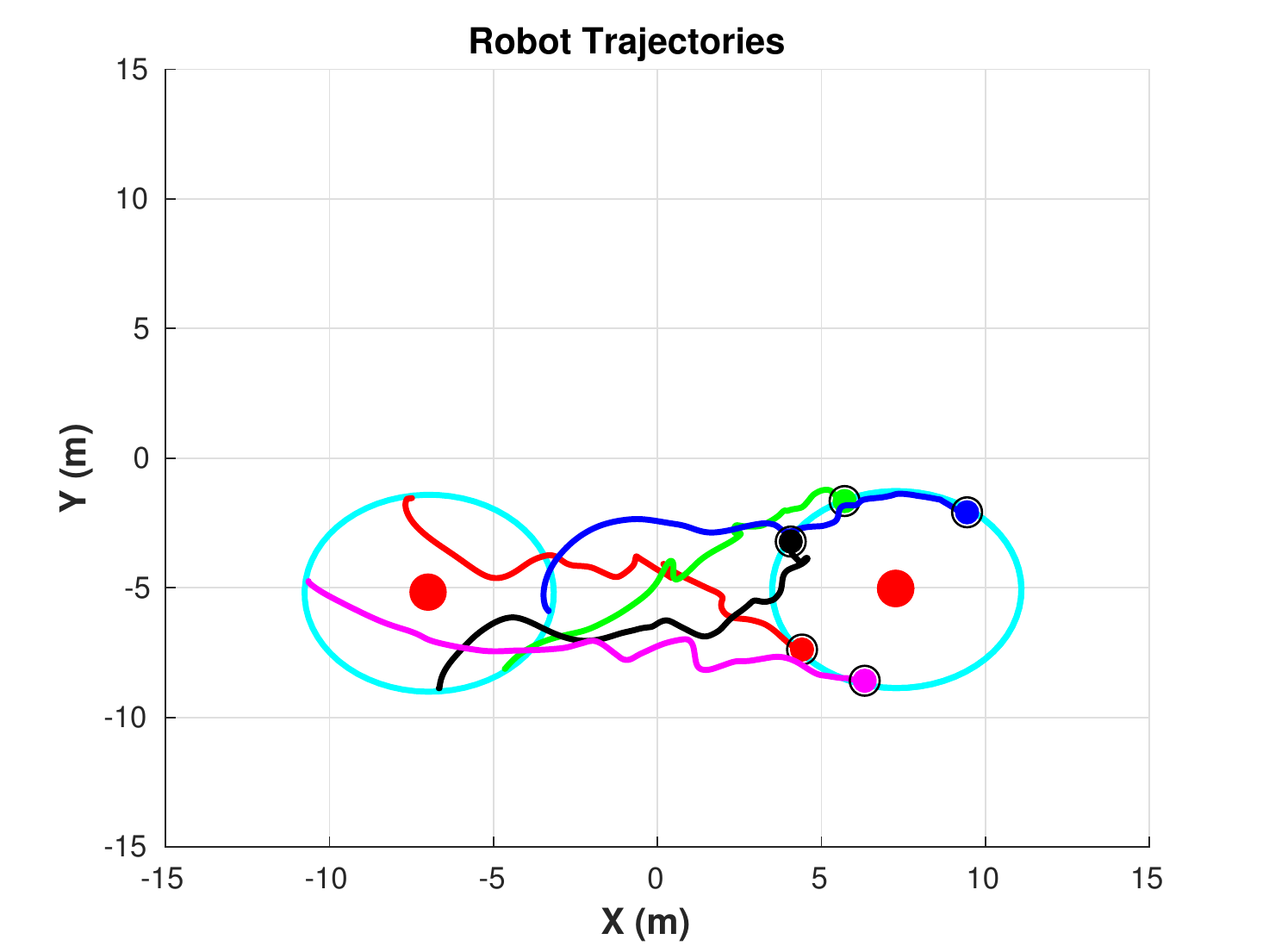}   
		\caption{}     
		\label{Swivel_5}
	\end{subfigure}
	\begin{subfigure}[t]{0.24\textwidth}        
		\includegraphics[scale=0.22]{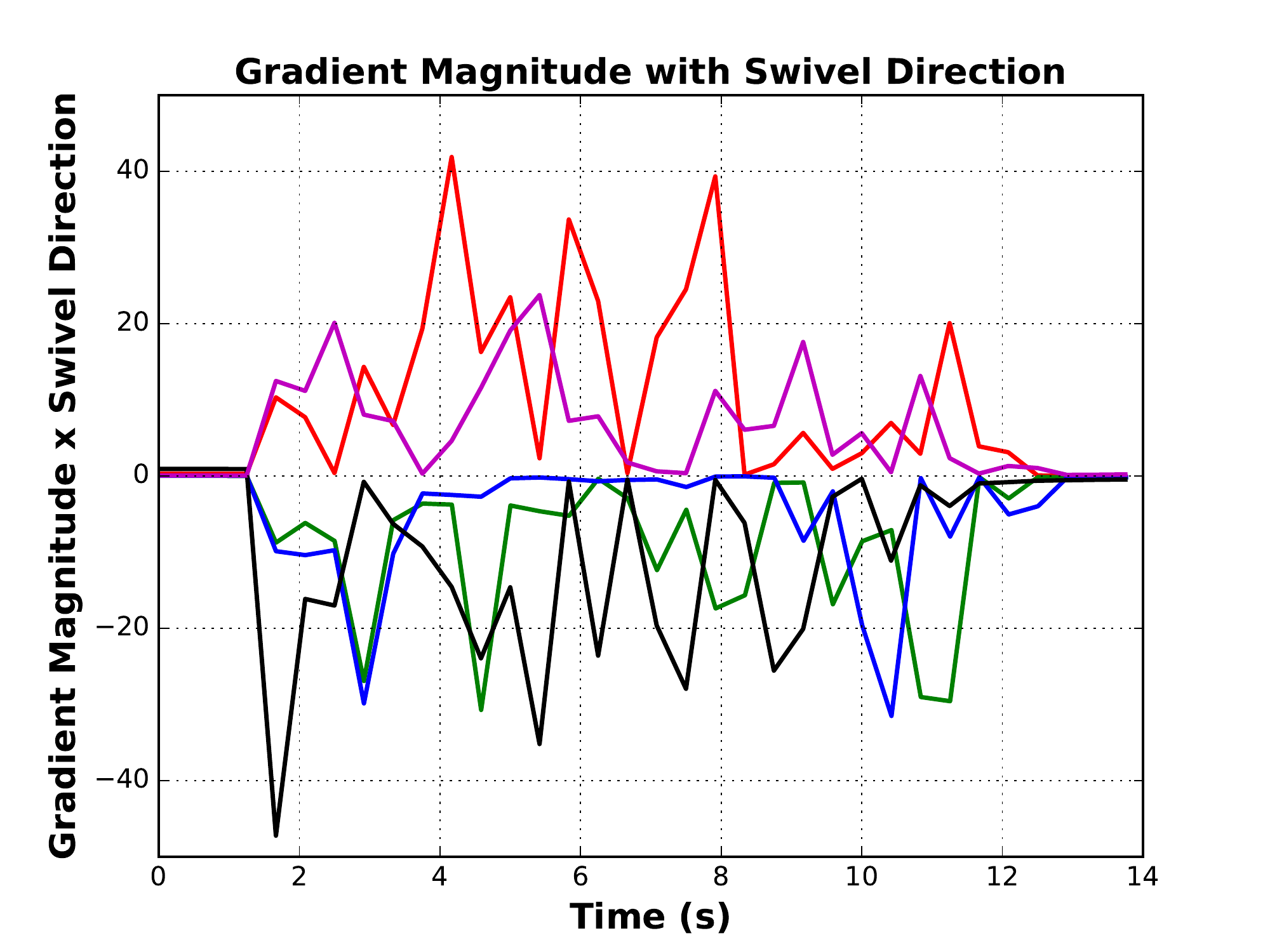}
		\caption{}
		\vspace{0.1cm}
		\label{Swivel_5_grad}
	\end{subfigure} 
	\begin{subfigure}[t]{0.24\textwidth}
		\centering
		\includegraphics[scale=0.27]{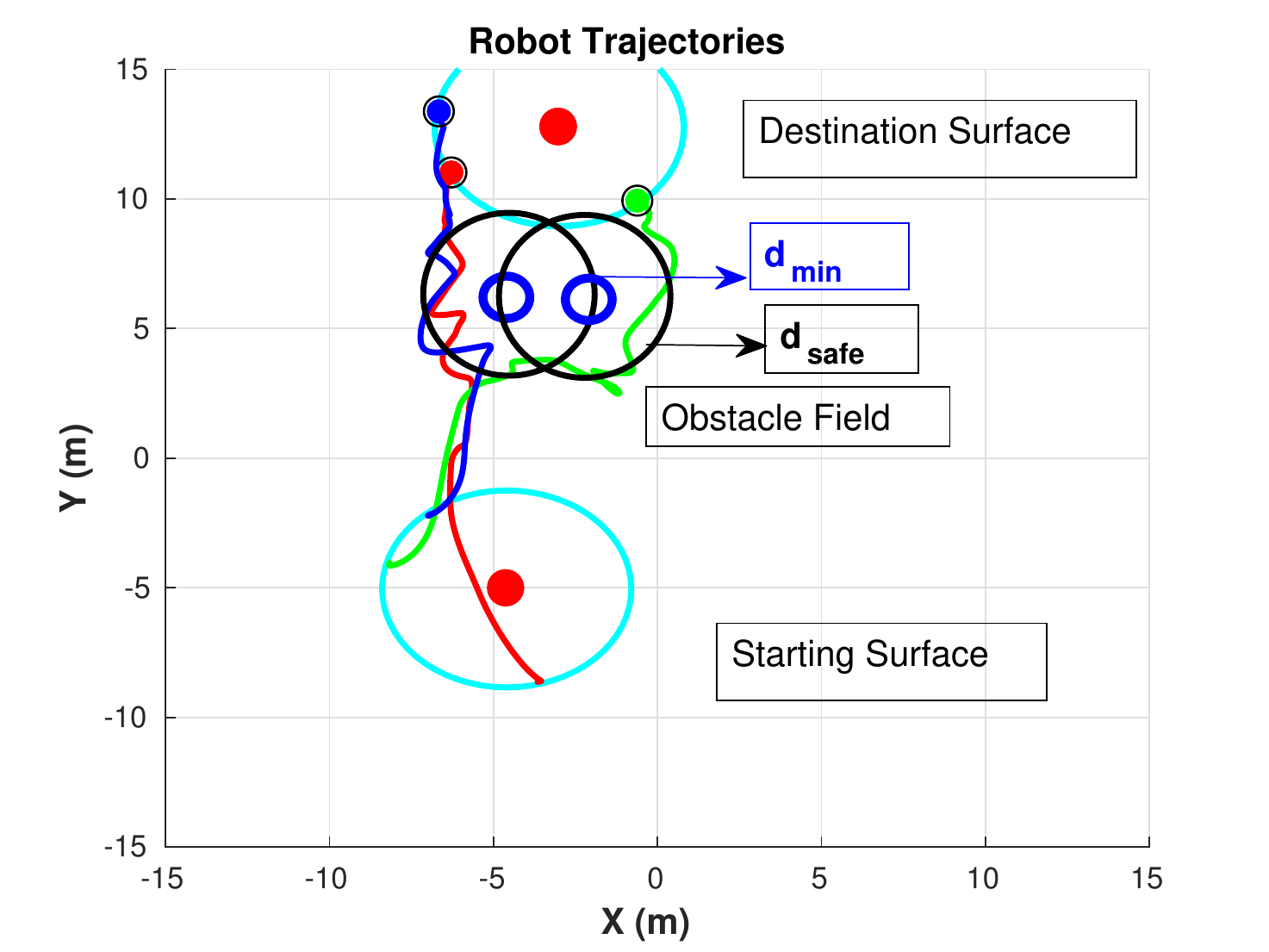}  
		\caption{}    
		\label{Swivel_32}
	\end{subfigure}
	\begin{subfigure}[t]{0.24\textwidth}
		\centering
		\includegraphics[scale=0.21]{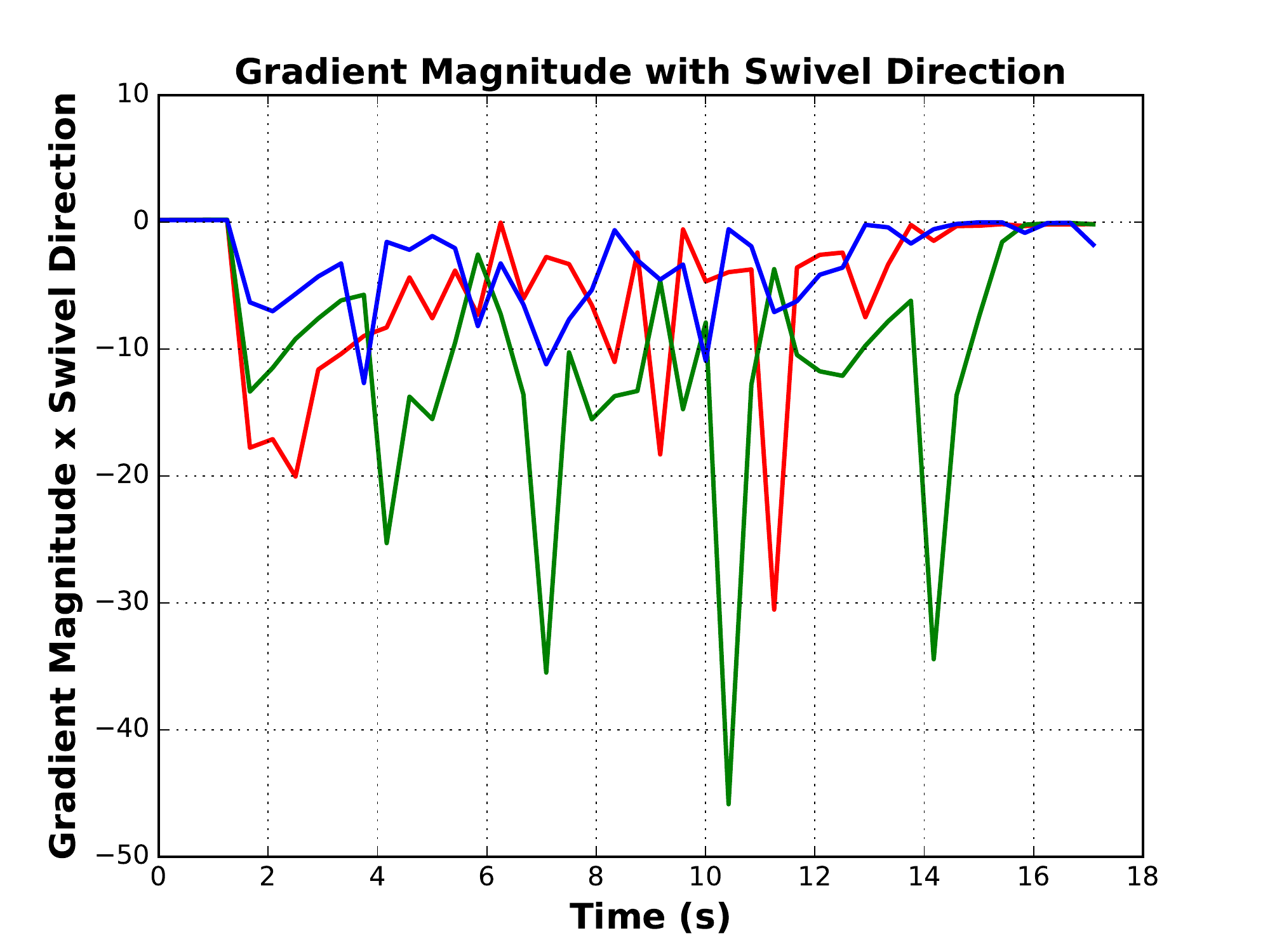}
		\caption{}
		\vspace{0.1cm}
		\label{Swivel_32_grad}
	\end{subfigure}
	\begin{subfigure}[t]{0.24\textwidth}
		\centering
		\includegraphics[scale=0.27]{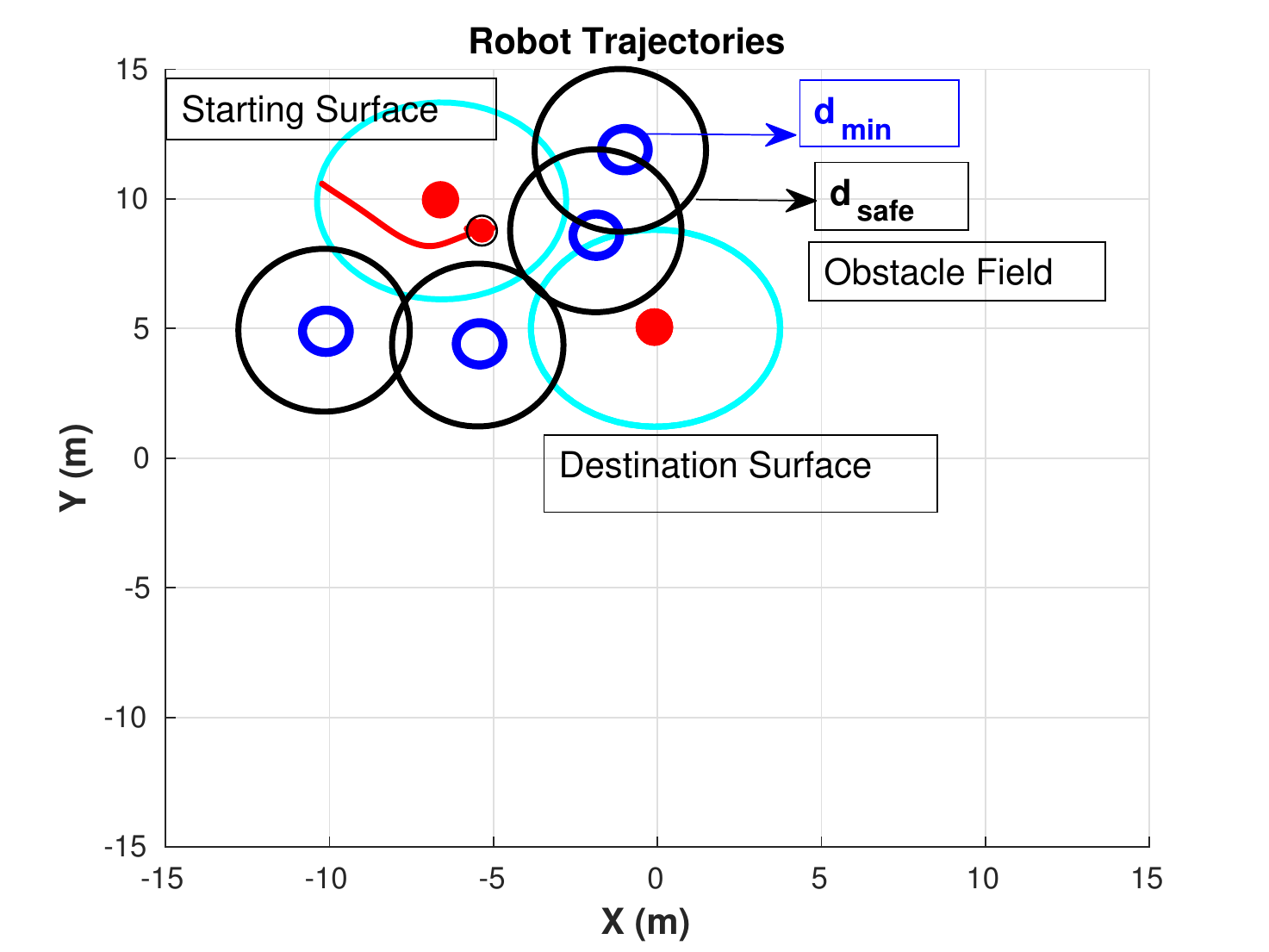} 
		\caption{}   	
		\label{Swivel_14}
	\end{subfigure}
	\begin{subfigure}[t]{0.24\textwidth}
		\centering
		\includegraphics[scale=0.22]{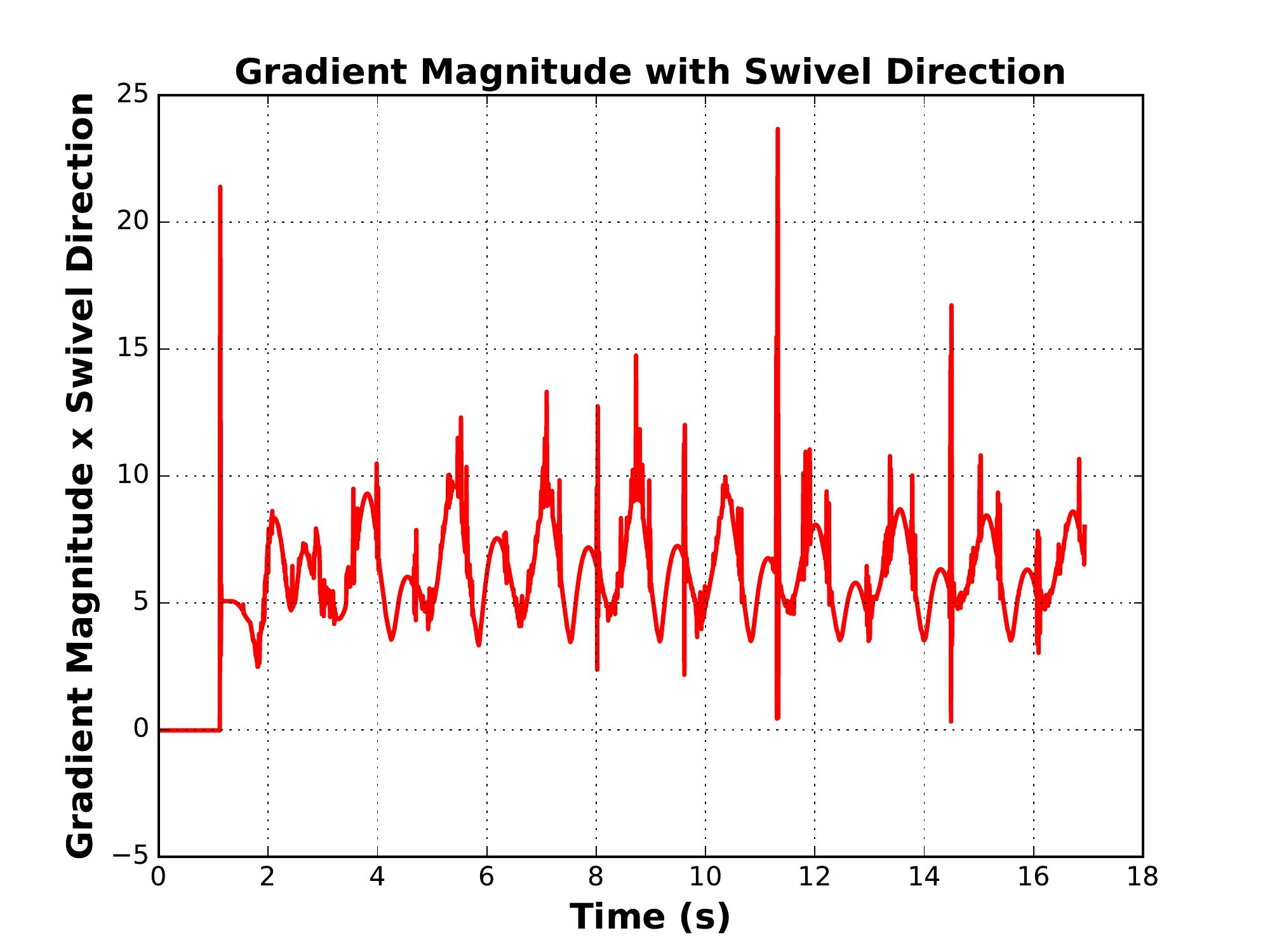}
		\caption{}
		\label{Swivel_14_grad}
	\end{subfigure}      
    \caption{MAV trajectories and optimization gradients of swivelling destination method.}
    \label{Swivel_Result} 
    \vspace{-1em}
\end{figure}
\vspace{-0.5em}
\subsection{\label{sub:expsetup}Experimental Setup}
The algorithms  were simulated in a Gazebo+ROS integrated environment to emulate the real world physics and enable a decentralized implementation for validation of the proposed methods. The setup runs on a standalone Intel Xeon E5-2650 processor. The simulation environment consists of multiple hexarotor MAVs confined in a 3D world of $20 m\times 20 m\times 20 m$. All the experiments were conducted using multiple MAVs for 3 different task scenarios involving simultaneous target tracking and obstacle avoidance namely,
\begin{itemize}
\item Scenario I: 5 MAVs need to traverse from a starting surface to destination surface without collision. Each agent acts as a dynamic obstacle to every other MAV. 
\item Scenario II: 2 MAVs hover at certain height and act as static obstacles. The remaining 3 MAVs need to reach the desired surface avoiding static and dynamic obstacles.
\item Scenario III: 4 MAVs hover and form a U-shaped static obstacle. The remaining MAV must reach destination while avoiding field local minima and control deadlock.
\end{itemize}
It may be noted that in all the above scenarios, the target's position is drastically changed from  initial to final destination. This is done so as to create a more challenging target tracking scenario than simple target position transitions.
Furthermore, the scalability and effectiveness of the proposed algorithms are verified by antipodal position swap of 8 MAVs within a surface. The MAVs perform 3D obstacle avoidance to reach their respective positions while ensuring that the surface center (target) is always in sight.

The convex optimization \eqref{DQMPC}-\eqref{last_DQMPC} is solved as a quadratic program using CVXGEN \cite{mattingley2012cvxgen}. The DQMPC operates at a rate of $100$ Hz. The state and velocity limits of each MAV $R_k$ are $[-20,-20,3]^{\top} \leq ~ \x_t^{R_k}(n) \leq ~ [20,20,10] ^{\top} $ in $m$ and $[-5,-5,-5]^{\top} \leq ~ \dot{\x}_t^{R_k}(n) \leq ~ [5,5,5]^{\top} $ in $m/s$ respectively, while the control limits are  $[-2,-2,-2]^{\top} \leq \ao_t^{R_k}(n) \leq [2,2,2]$ in $m/s^2$. 
The desired hovering height of each MAV is $h_{gnd} = 5 m$  and the yaw $\psi^{R_k}$ of each MAV is oriented towards the target. The horizon $N$ for the DQMPC and potential force computation is $15$ time steps each. It is important to mention that if no trajectory information is available for an obstacle or adversary $O_j$, the $n=0$ magnitude of potential field ($F_{rep}^{R_k,O_j}(0)$) is used for the entire horizon. $d_{min}= 0.5m$ and $d_{safe}= 3m$ for the potential field around obstacles. The destination surface is circular, with radius $d^{R_k} = 4 m ~ \forall k$,  around the target for all experiments. However, as stated earlier our approach can attain any desired 3D destination surface.

\subsection{DQMPC: Baseline Method}
Figure \ref{DQMPC_Result} showcases the multi-robot target tracking results for the three different scenarios (see Sec. \ref{sub:expsetup}) while applying the baseline DQMPC method. As clearly seen in Fig. \ref{DQMPC_Result}(a,c), the agents find obstacle free trajectories from starting to destination surface for the scenarios I and II. This is also indicated by the magnitude of gradient dropping close to $0$ after $40s$ and $30s$ respectively as seen in Fig. \ref{DQMPC_Result}(b,d). The rapidly varying gradient curve of Scenario I (Fig. \ref{DQMPC_Result}(b)) shows that, each agent's potential field pushes other agents to reach the destination surface\footnote{Note that the scale of the gradient plots are different. The magnitude of the gradient is just to suggest that the robot has reached the target surface.}. We observe that the MAVs converge to the destination (Fig. \ref{DQMPC_Result}(c)) by avoiding the obstacle fields. 

In the U-shaped static obstacle case (scenario III), the agent is stuck because of control deadlock and fails to reach the destination surface as seen in Fig. \ref{DQMPC_Result}(e). The periodic pattern of the gradient curve (Fig. \ref{DQMPC_Result}(f)) and  non-zero gradient magnitude makes the deadlock situation evident. Most of these scenarios can also be visualized in the attached video submission.
\begin{figure}
	\centering
	\begin{subfigure}[t]{0.24\textwidth}        
		\includegraphics[scale=0.27]{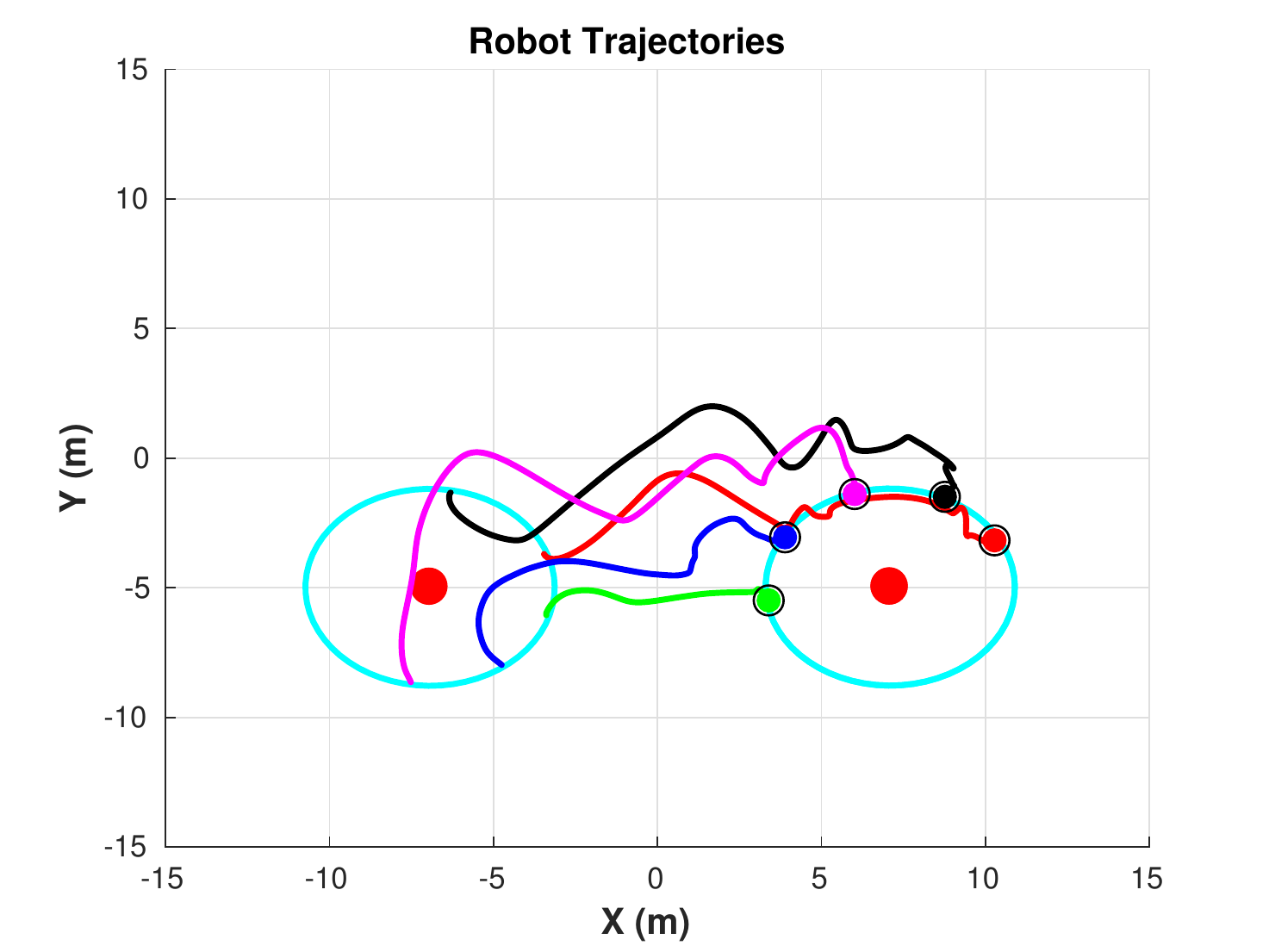}   
		\caption{}     
		\label{Ang_5}
	\end{subfigure}
	\begin{subfigure}[t]{0.24\textwidth}        
		\includegraphics[scale=0.22]{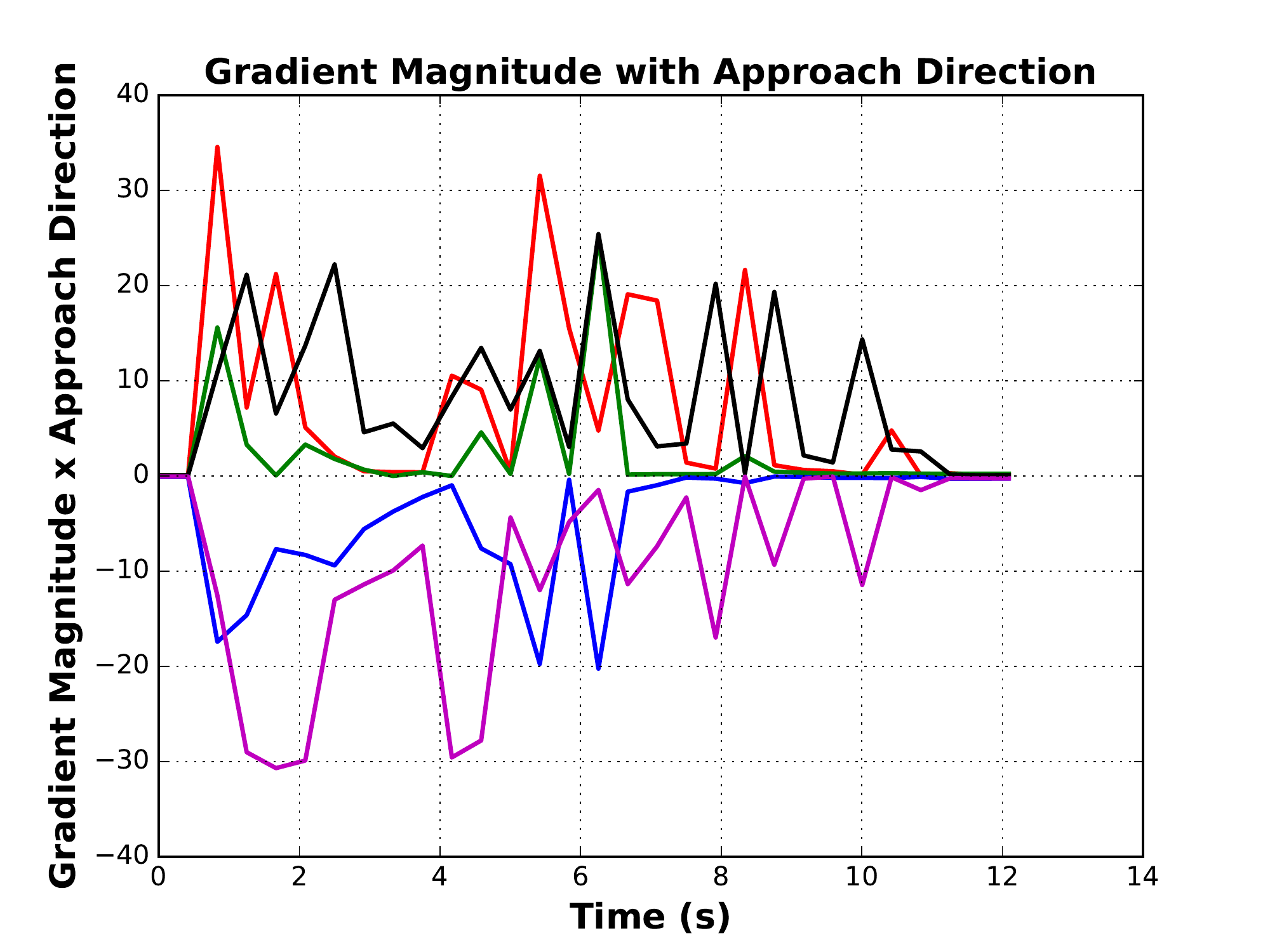}
		\caption{}
		\vspace{0.1cm}
		\label{Ang_5_grad}
	\end{subfigure} 
	\begin{subfigure}[t]{0.24\textwidth}
		\centering
		\includegraphics[scale=0.27]{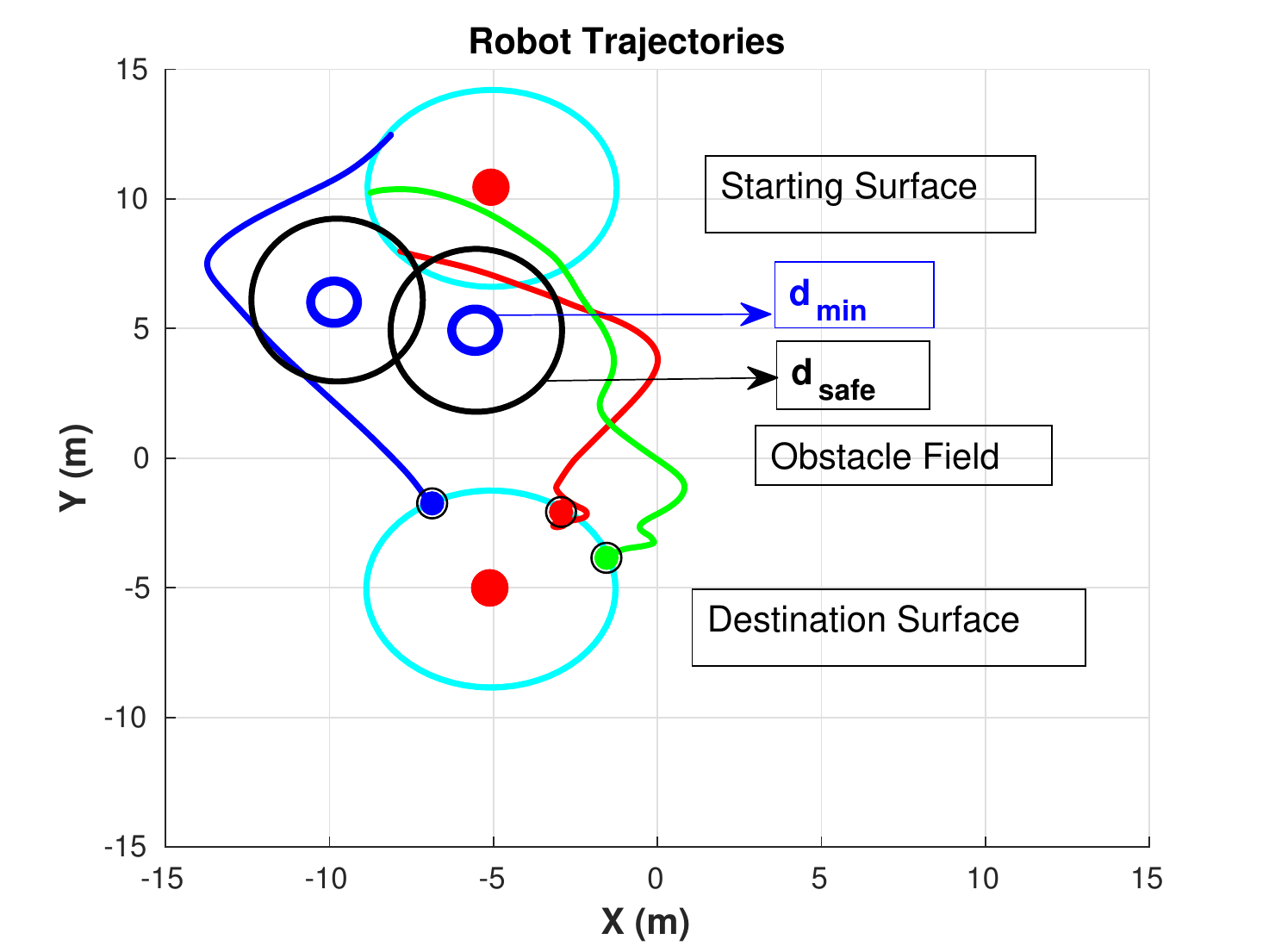}  
		\caption{}    
		\label{Ang_32}
	\end{subfigure}
	\begin{subfigure}[t]{0.24\textwidth}
		\centering
		\includegraphics[scale=0.21]{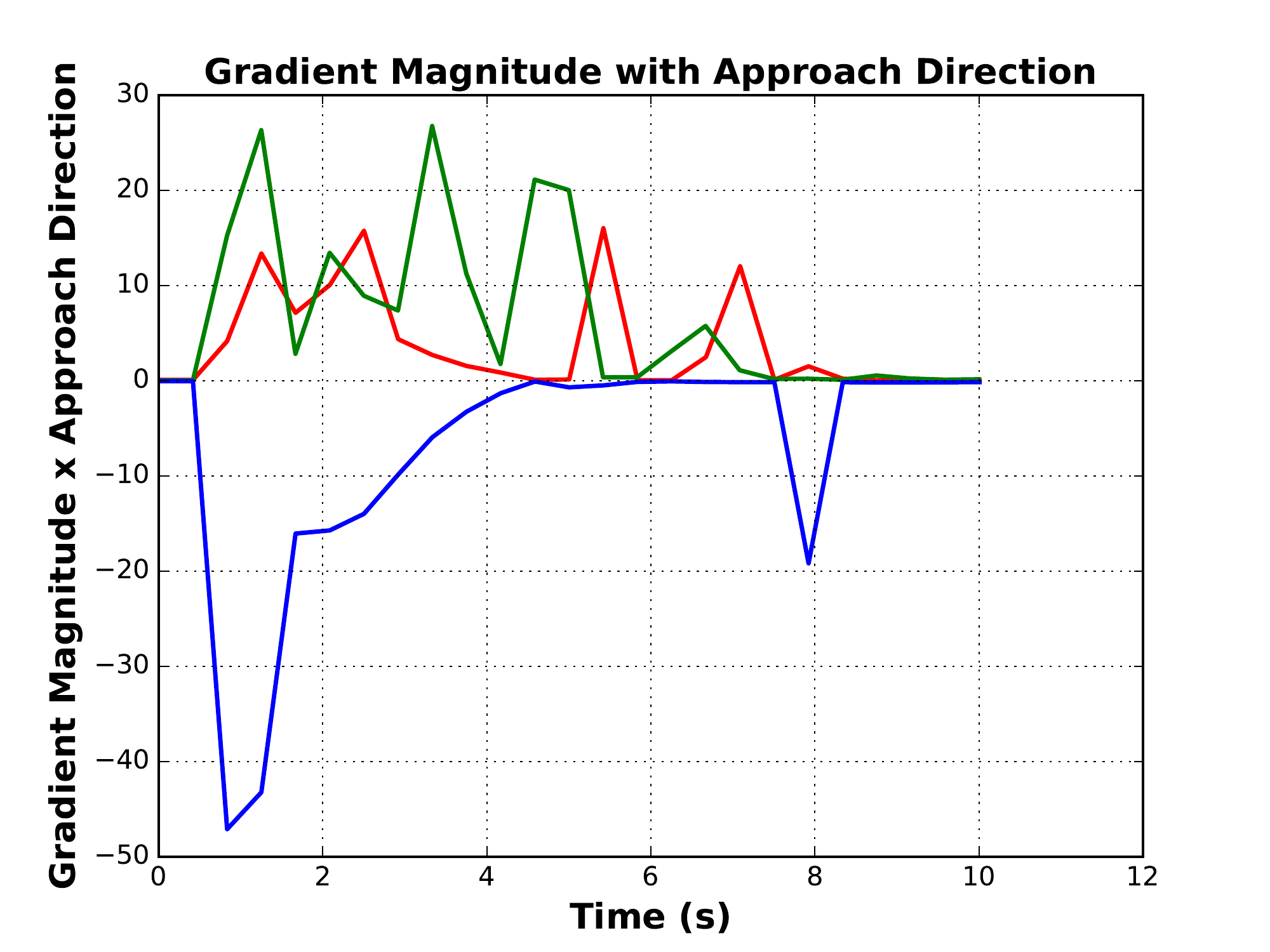}
		\caption{}
		\vspace{0.1cm}
		\label{Ang_32_grad}
	\end{subfigure}
	\begin{subfigure}[t]{0.24\textwidth}
		\centering
		\includegraphics[scale=0.27]{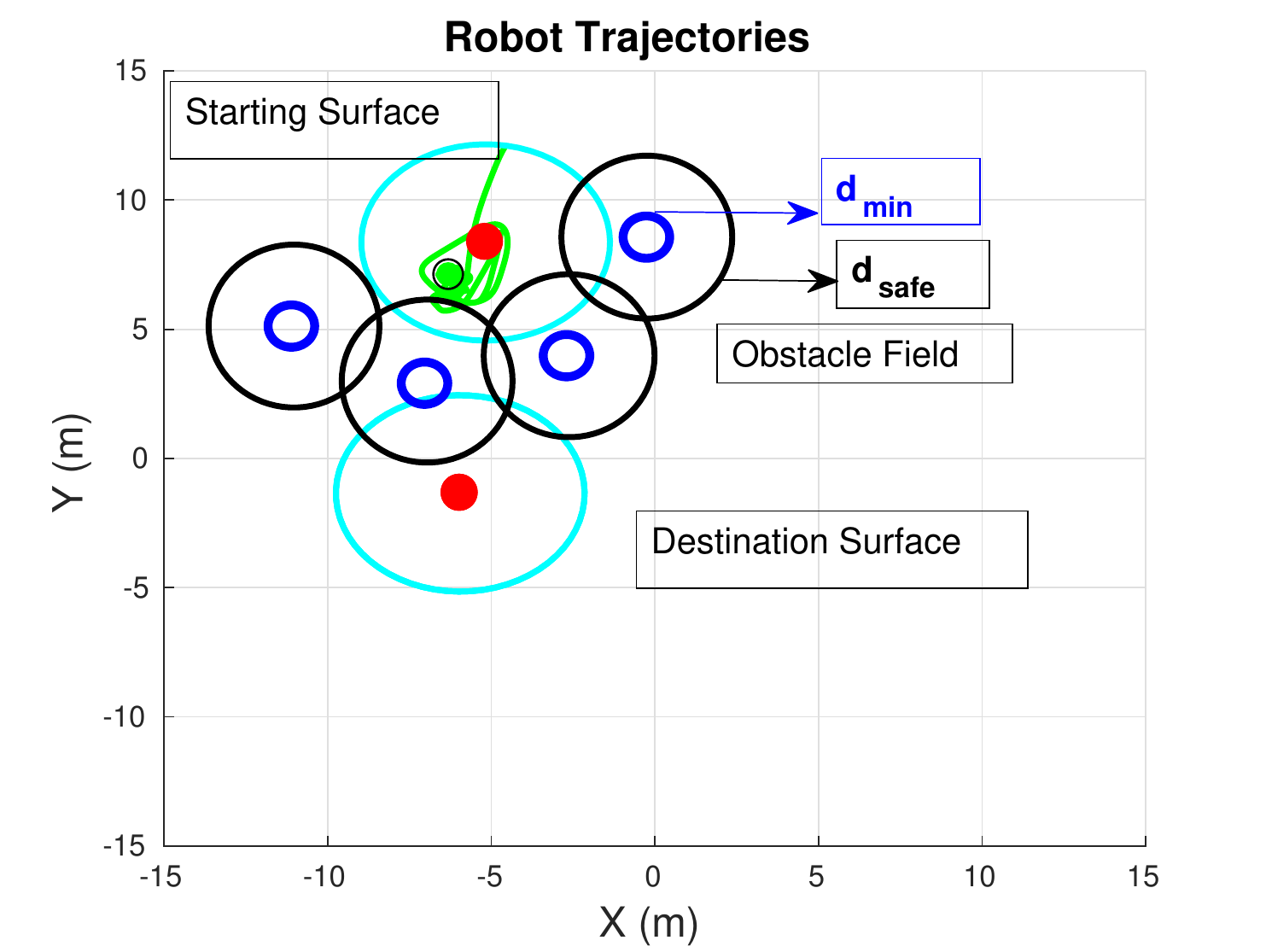} 
		\caption{}   	
		\label{Ang_14}
	\end{subfigure}
	\begin{subfigure}[t]{0.24\textwidth}
		\centering
		\includegraphics[scale=0.22]{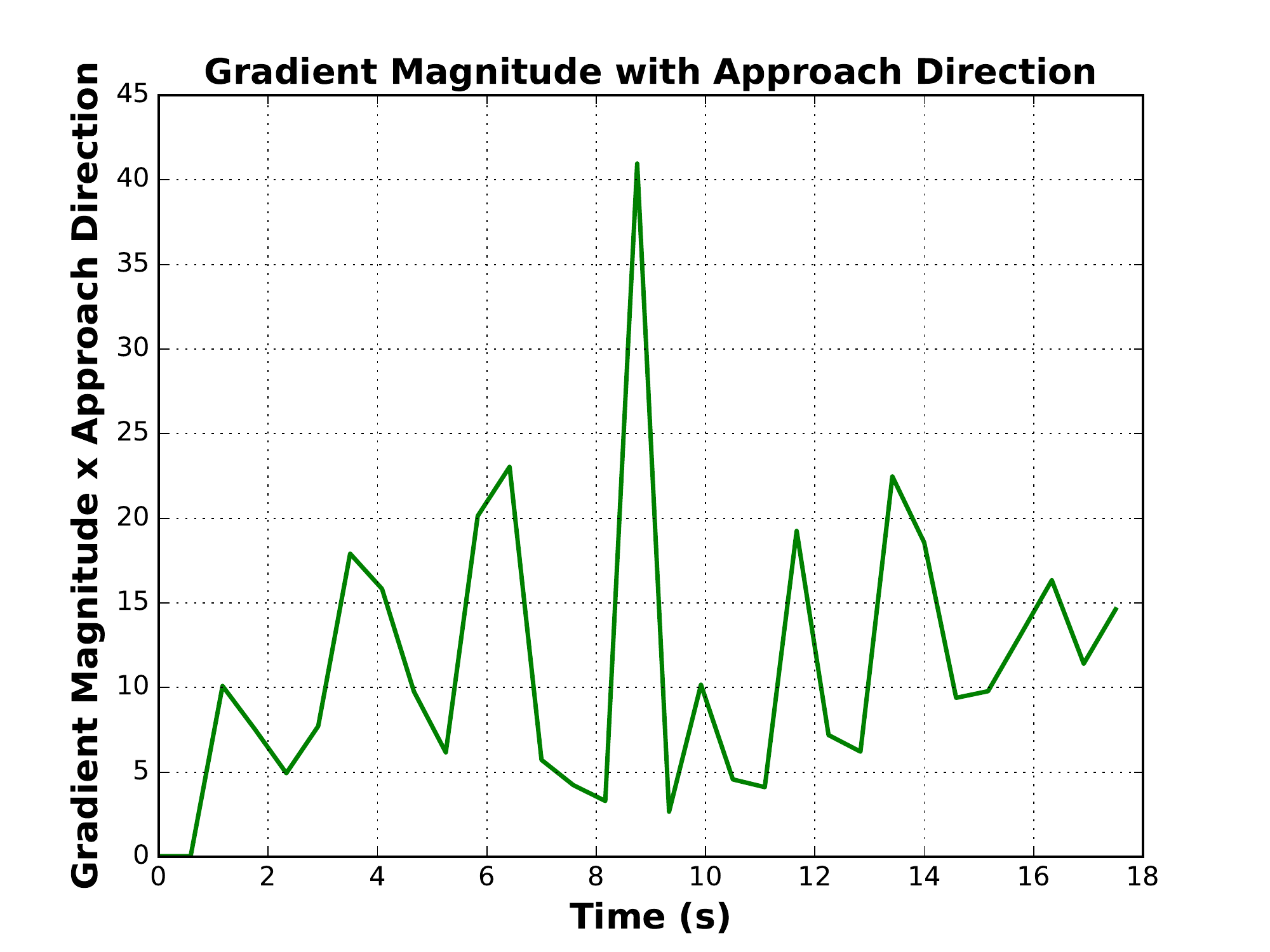}
		\caption{}
		\label{Ang_14_grad}
	\end{subfigure}      
	\caption{MAV trajectories and optimization gradients of approach angle force.}
	\vspace{-1em}
	\label{Ang_Result} 
\end{figure}
%

\subsection{Swivelling Robot Destination Method}
In the swivelling destination method, we use $k_s = 0.05$ as the gain for gradient magnitude impact. Fig. \ref{Swivel_Result}(a) shows, the MAVs spreading themselves along the destination surface depending on their distance from target and therefore the gradient. This method has a good convergence time of about $12s$ and $15s$ (about 3 times faster than the baseline DQMPC) for the scenarios I and II as can be observed in  Fig. \ref{Swivel_Result}(b) and  Fig. \ref{Swivel_Result}(d) respectively. Here, the direction of swivel depends on the agents orientation w.r.t. to target and each agent takes a clockwise or anti-clockwise swivel based on its orientation as clearly seen in  Fig. \ref{Swivel_Result}(c). The positive and negative values of the gradient indicate the direction. It can be observed in  Fig. \ref{Swivel_Result}(c), that the agent at times crosses over the $d_{safe}$ because of the higher target attraction force but respects the $d_{min}$ where the repulsion force is infinite.
Despite the better convergence time, the MAV is stuck when encountered with a U-shaped static obstacle (Fig. \ref{Swivel_Result}(e)) because of a control deadlock. This can also be observed with the non-zero gradient in Fig. \ref{Swivel_Result}(f).

\subsection{Approach Angle Towards Target Method}
Figure \ref{Ang_Result} showcases the results of the approach angle method. Since the additional potential field force enforces that not more than one agent has the same approach angle, the MAVs spread themselves while approaching the destination surface. This is associated with fast convergence to target surface as observed in Fig. \ref{Ang_Result}(a,b,c,d), for the scenarios I and II. The direction of the approach angle depends on the orientation of the MAVs w.r.t. to the target and is also indicated by the positive and negative values of the gradient magnitude. From the many experiments conducted in environments having only dynamic obstacles or sparsely spaced static obstacles, we observed that this method has the smoothest transition to the destination surface. However, similar to the swivelling destination method, it also fails to overcome control deadlock in case of a U-shaped obstacle as observed in Fig. \ref{Ang_Result}(e,f).

\begin{figure}
	\centering
	\begin{subfigure}[t]{0.24\textwidth}        
		\includegraphics[scale=0.27]{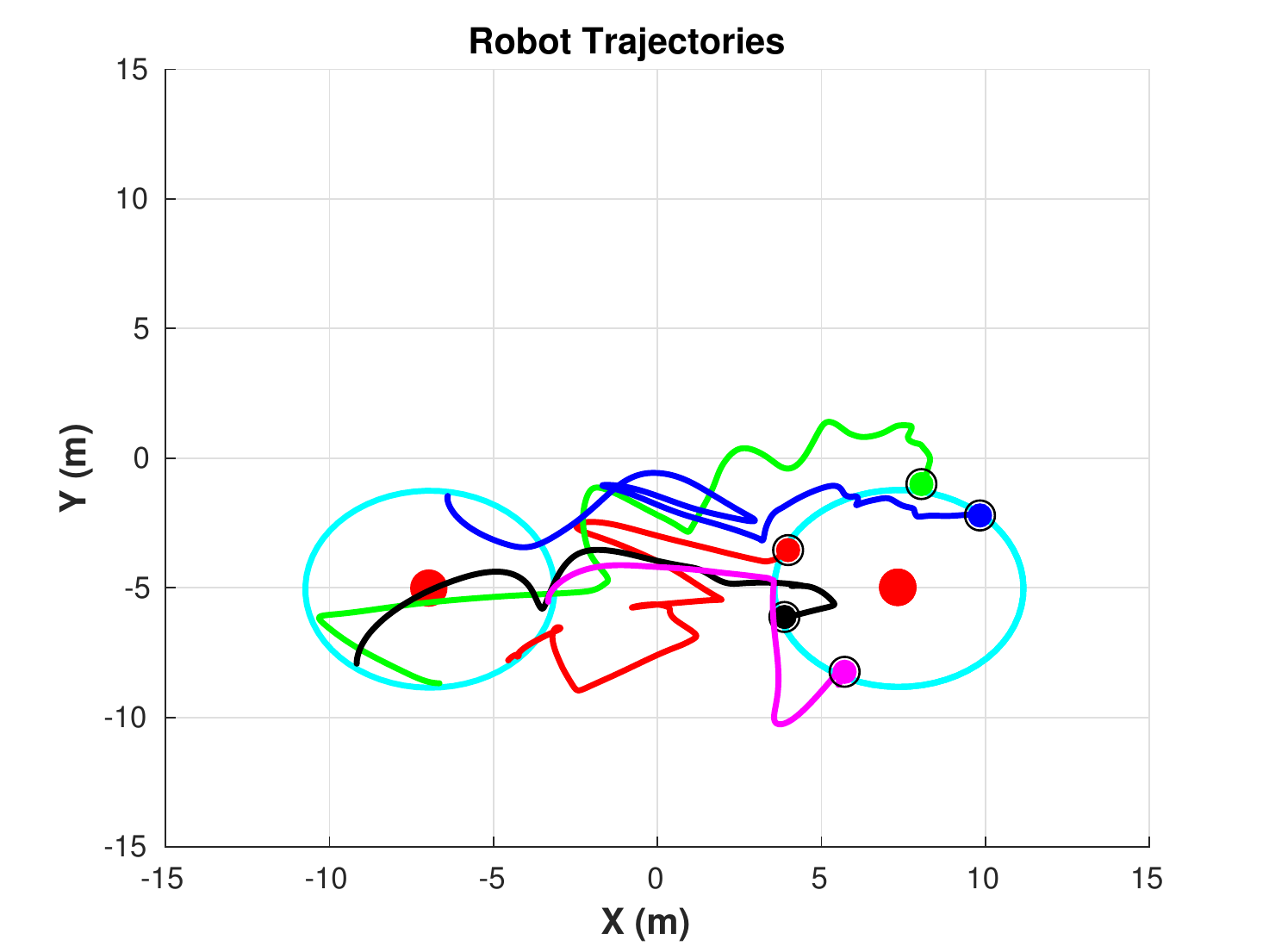}   
		\caption{}     
		\label{Tang_5}
	\end{subfigure}
	\begin{subfigure}[t]{0.24\textwidth}        
		\includegraphics[scale=0.22]{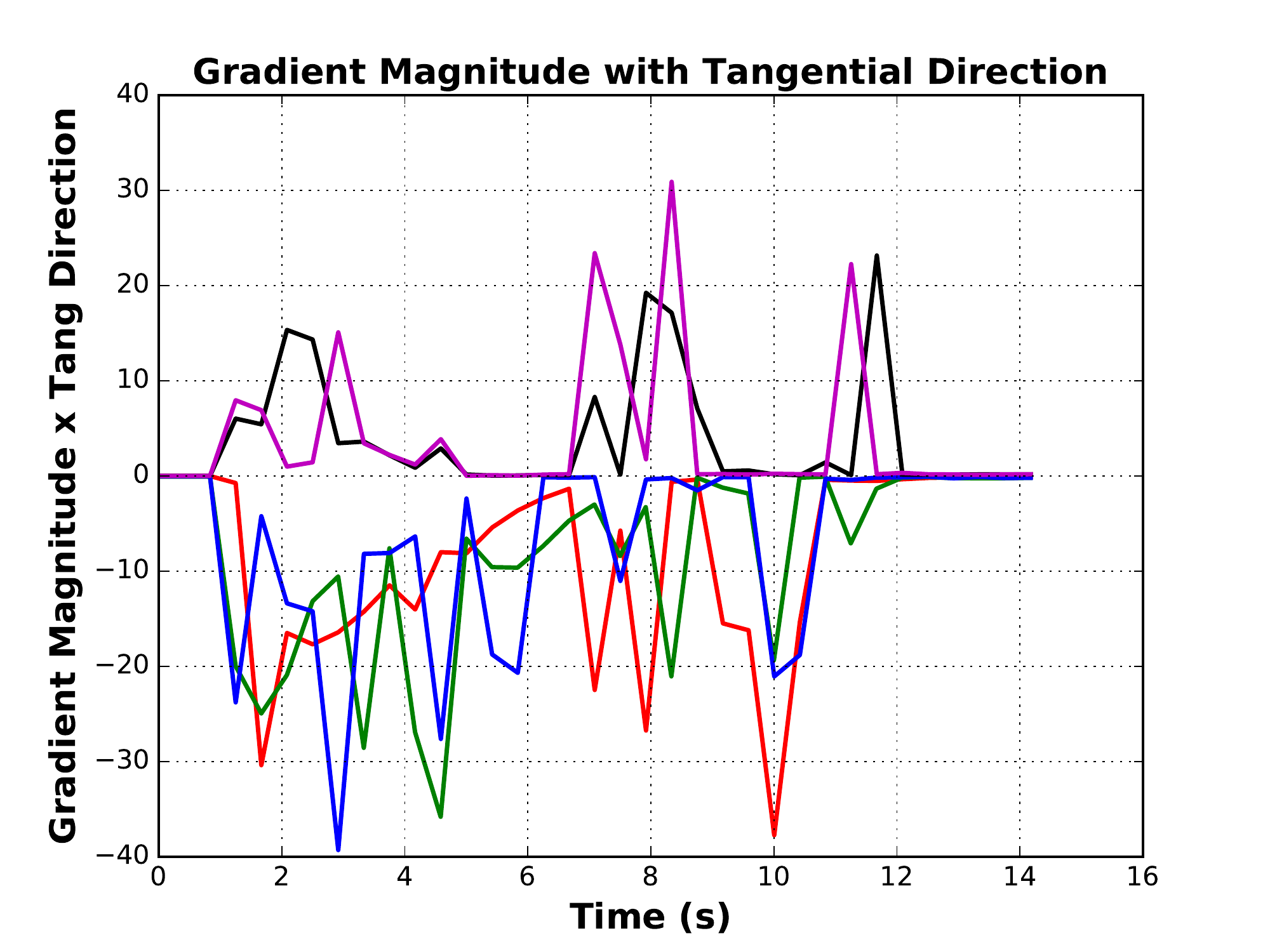}
		\caption{}
		\vspace{0.1cm}
		\label{Tang_5_grad}
	\end{subfigure} 
	\begin{subfigure}[t]{0.24\textwidth}
		\centering
		\includegraphics[scale=0.27]{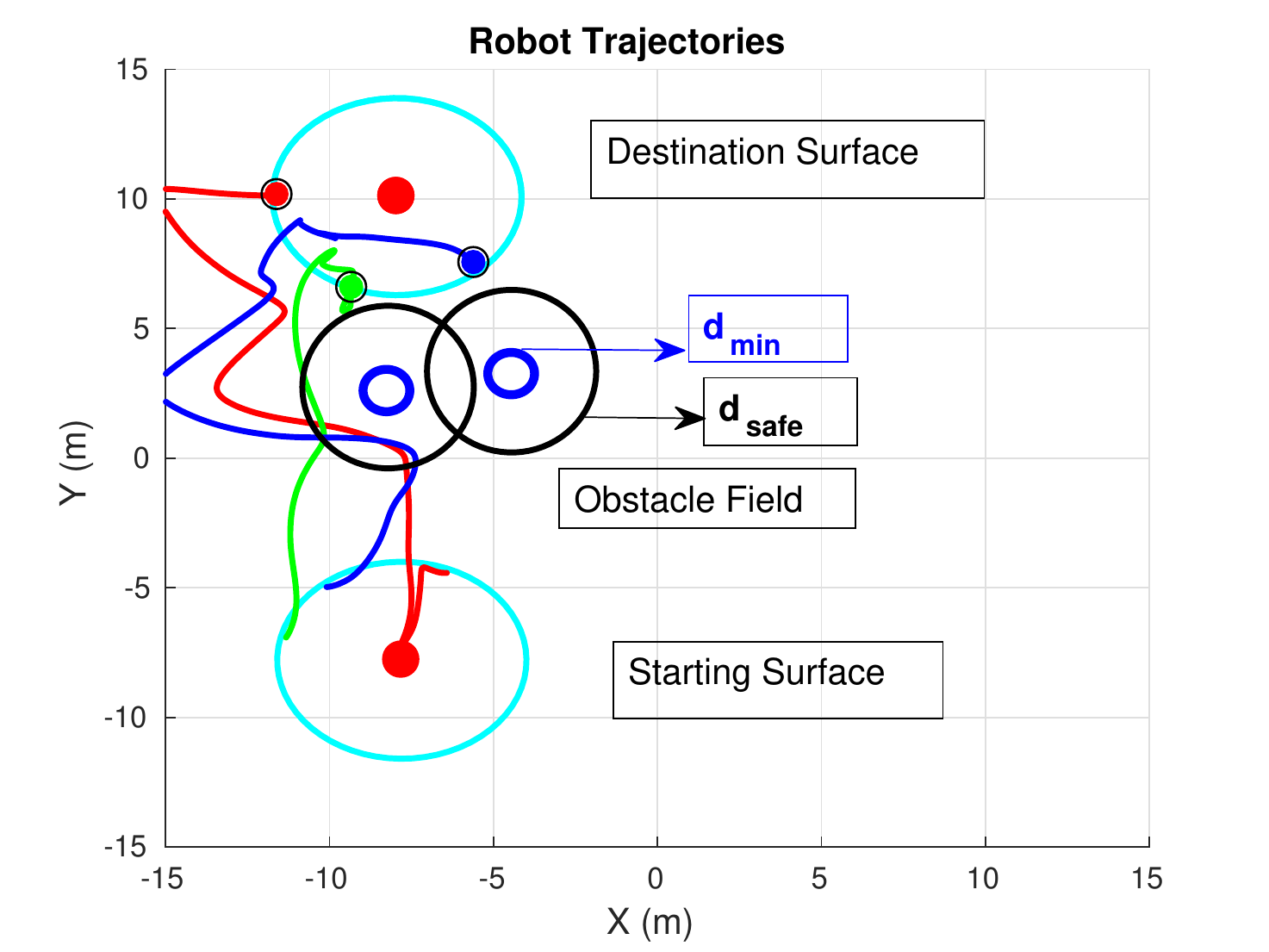}  
		\caption{}    
		\label{Tang_32}
	\end{subfigure}
	\begin{subfigure}[t]{0.24\textwidth}
		\centering
		\includegraphics[scale=0.21]{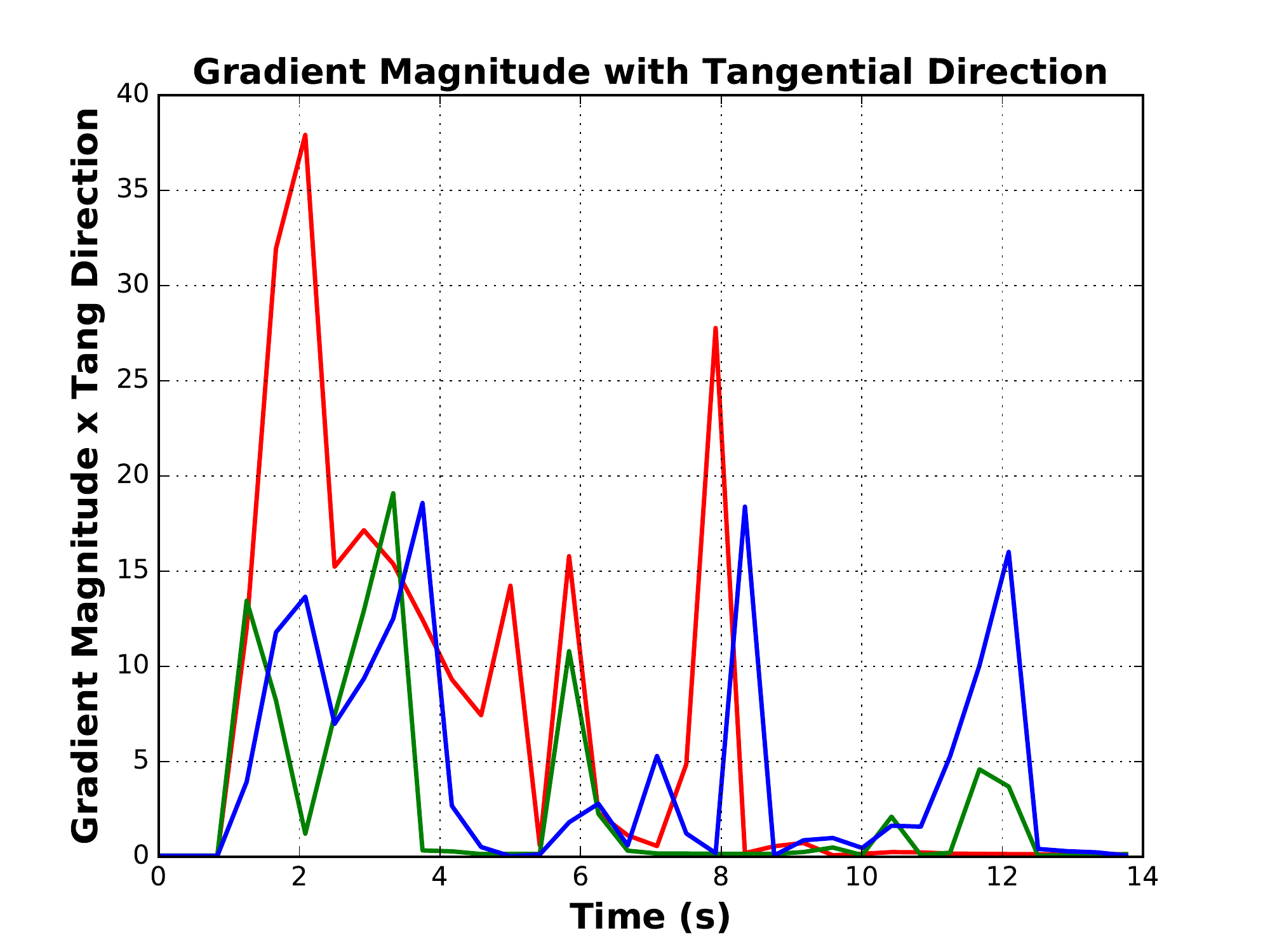}
		\caption{}
		\vspace{0.1cm}
		\label{Tang_32_grad}
	\end{subfigure}
	\begin{subfigure}[t]{0.24\textwidth}
		\centering
		\includegraphics[scale=0.27]{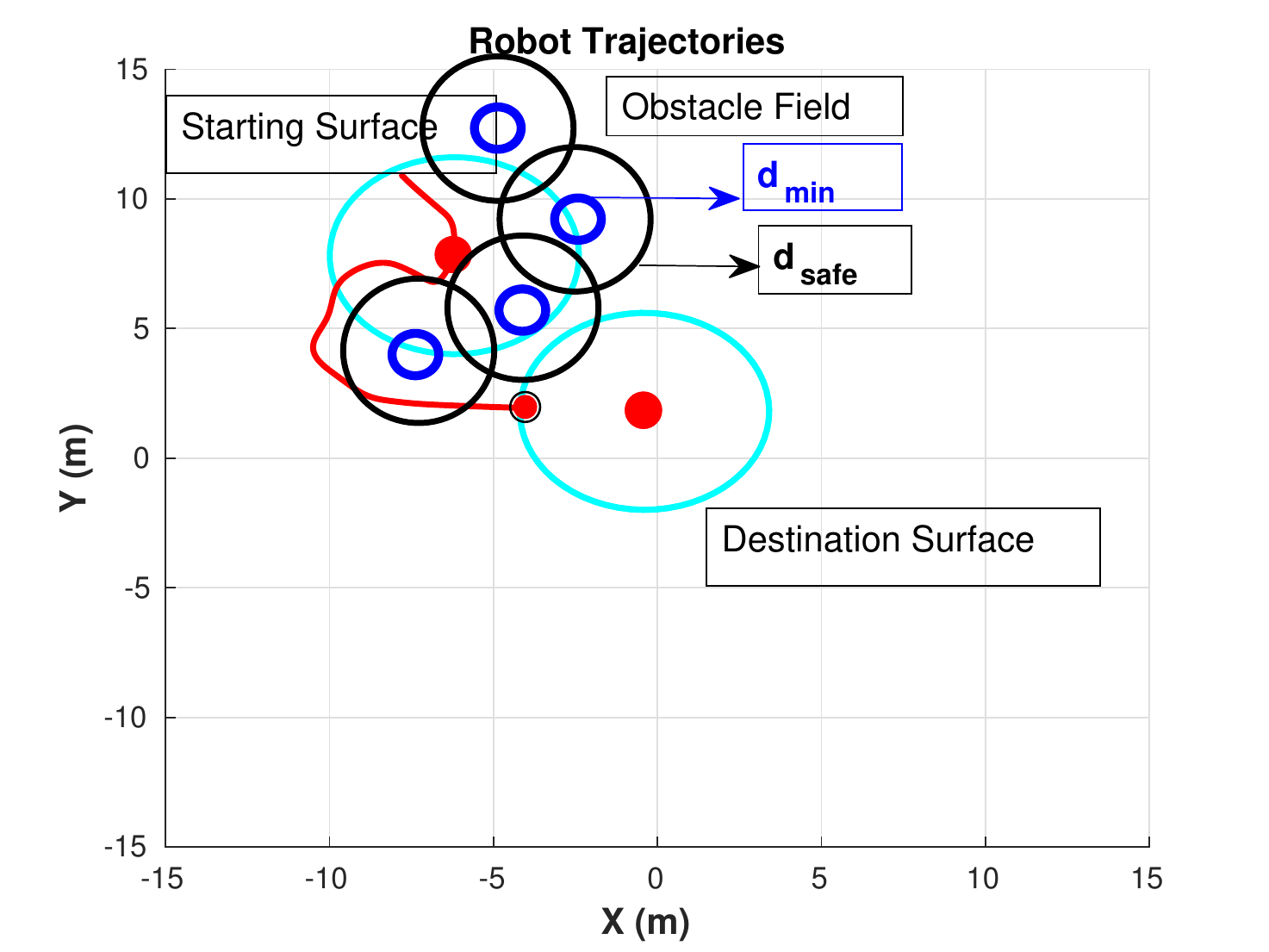} 
		\caption{}   	
		\label{Tang_14}
	\end{subfigure}
	\begin{subfigure}[t]{0.24\textwidth}
		\centering
		\includegraphics[scale=0.22]{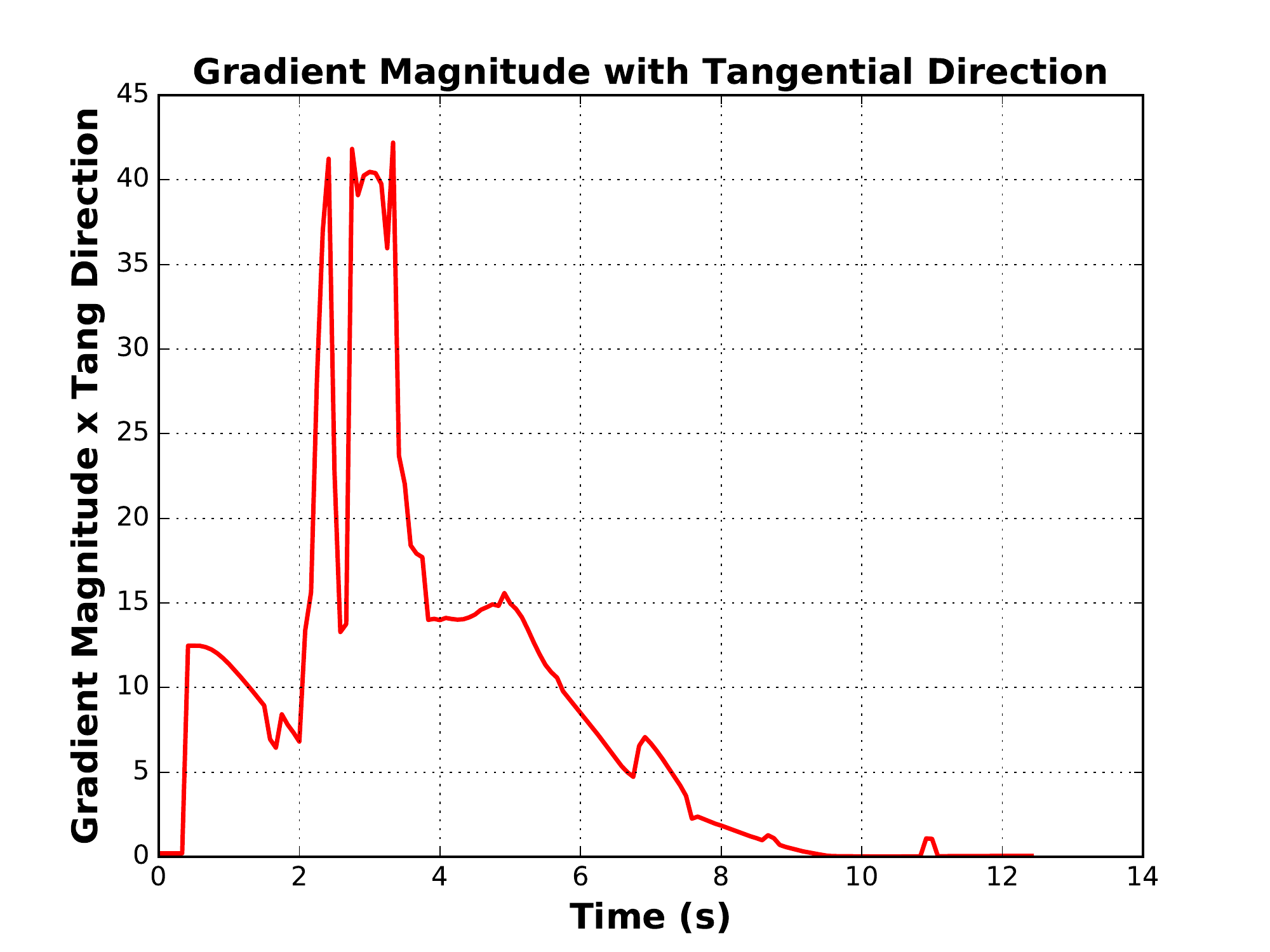}
		\caption{}
		\label{Tang_14_grad}
	\end{subfigure}      
	\caption{MAV trajectories and optimization gradients of tangential band method.}
	\label{Tang_Result}
\end{figure}

\subsection{Tangential Band Method}
The MAVs using tangential band method reach the destination surface in all the scenarios of static and dynamic obstacles as seen in Fig.~\ref{Tang_Result}. This method facilitates convergence to the target, for complex static obstacles, because, by principle the MAV traverses within the band until it finds a feasible path towards the target surface. $k_{tang}=2$  was used in simulations. As seen in Fig.~\ref{Tang_Result}(e), as soon as the MAV reaches obstacle surface, the tangential force acts, pushing it in the anti-clockwise direction. Then the UAV travels within this band until it is finally pulled towards the destination surface. The same principle applies for dynamic obstacles scenarios (see Fig.~\ref{Tang_Result}(a,c)) as well. Fig.~\ref{Tang_Result}(f) shows that, since the MAV overcomes field local minima and control deadlock, the gradient reaches $0$ for the U-shaped obstacle with a convergence time of $12$ seconds.

\begin{figure}[h]
\centering
\includegraphics[scale=0.32]{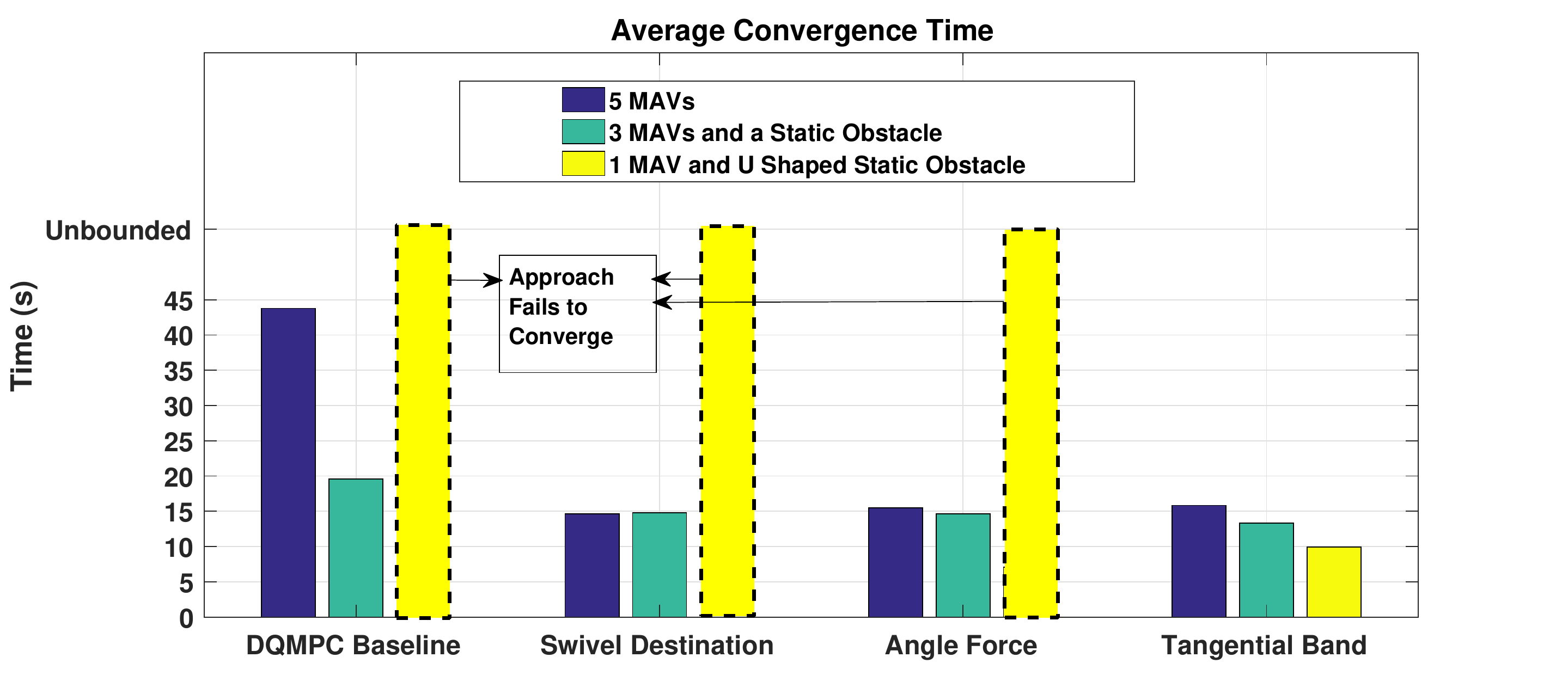}
\caption{Average convergence time comparison.}
\label{timing}
\vspace{-1em}
\end{figure}

\subsection{Convergence Time Comparison}
Figure \ref{timing} compares the average convergence time $T_{cvg}$ for each of the proposed methods for 3 trails, with approximately equal travel distances between starting and destination surface for the different scenarios mentioned in Sec.~\ref{sub:expsetup}. The $T_{cvg}$ for swivelling destination, approach angle and tangential band is approximately ($15s$), which is better compared to DQMPC in the scenarios of only dynamic (Scenario I) or simple static (Scenario II) obstacles. In the U-shaped obstacle (Scenario III), except tangential band the other methods get stuck in field local minima. It may be noted that all the method can be generalized for any shape of the destination surface. In summary, the tangential band method would be the most preferred choice when the type of obstacles in environment are unknown. All the mentioned results and additional experimental results can also be observed in the enclosed video file.

\subsection{Antipodal Movement}
In order to further emphasize the efficacy of tangential band and approach angle methods, we demonstrate  obstacle avoidance for a task of  intra-surface antipodal position swapping. Here 8 MAVs start on a circular surface with a radius of $8m$, at equal angular distance from each other w.r.t.\ the center (world frame origin $[0,0,0]^{\top}$). The MAVs must reach a point $180^{o}$ in the opposite direction while simultaneously maintaining their orientation towards the center. The plot in Tab. \ref{antipodal_approachangle_tangband} (a) shows the trajectories taken by MAVs in this task using the approach angle method. All the MAVs convergence to their antipodal positions in around $T_{cvg}=13s$. It may be noted that there is no trajectory specified and the MAVs compute their own optimal motion plans. To improve the convergence time in this scenario, a consistent clockwise direction is enforced, but this is not necessary and convergence is guaranteed either way. The video attachment visually showcases this antipodal movement.

Similarly the antipodal position swapping was carried out using the tangential band method with the same experimental criteria (see figure in Tab. \ref{antipodal_approachangle_tangband} (b)). A convergence time of around $T_{cvg}=11s$ was observed, further emphasizing the speed of obstacle avoidance and convergence of the proposed method.  

\begin{table}[t]
\begin{tabular}{cc}
 \centering
\hspace*{-10pt} \includegraphics[scale=0.32]{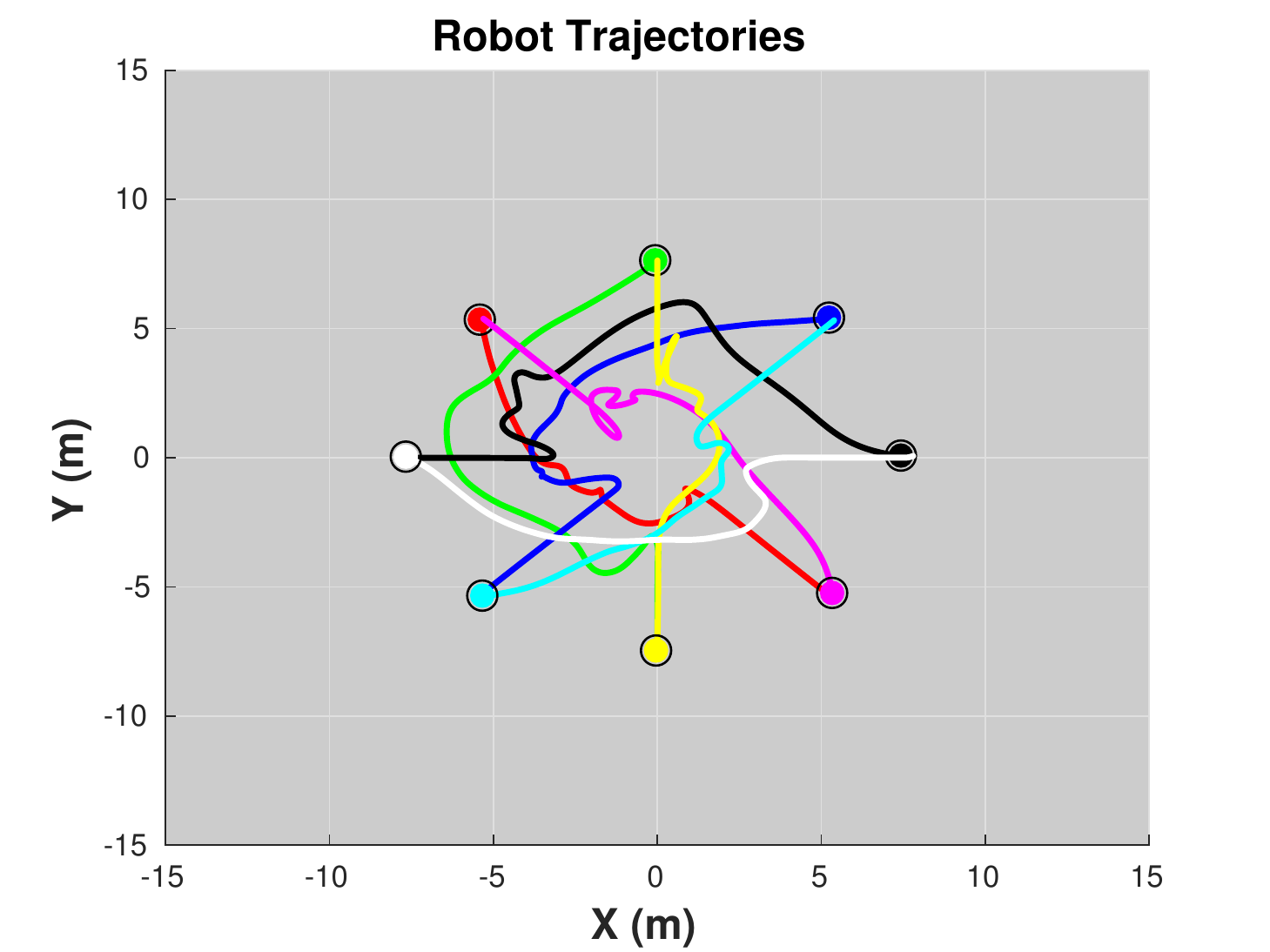} &  \hspace*{-28pt} \includegraphics[scale=0.32]{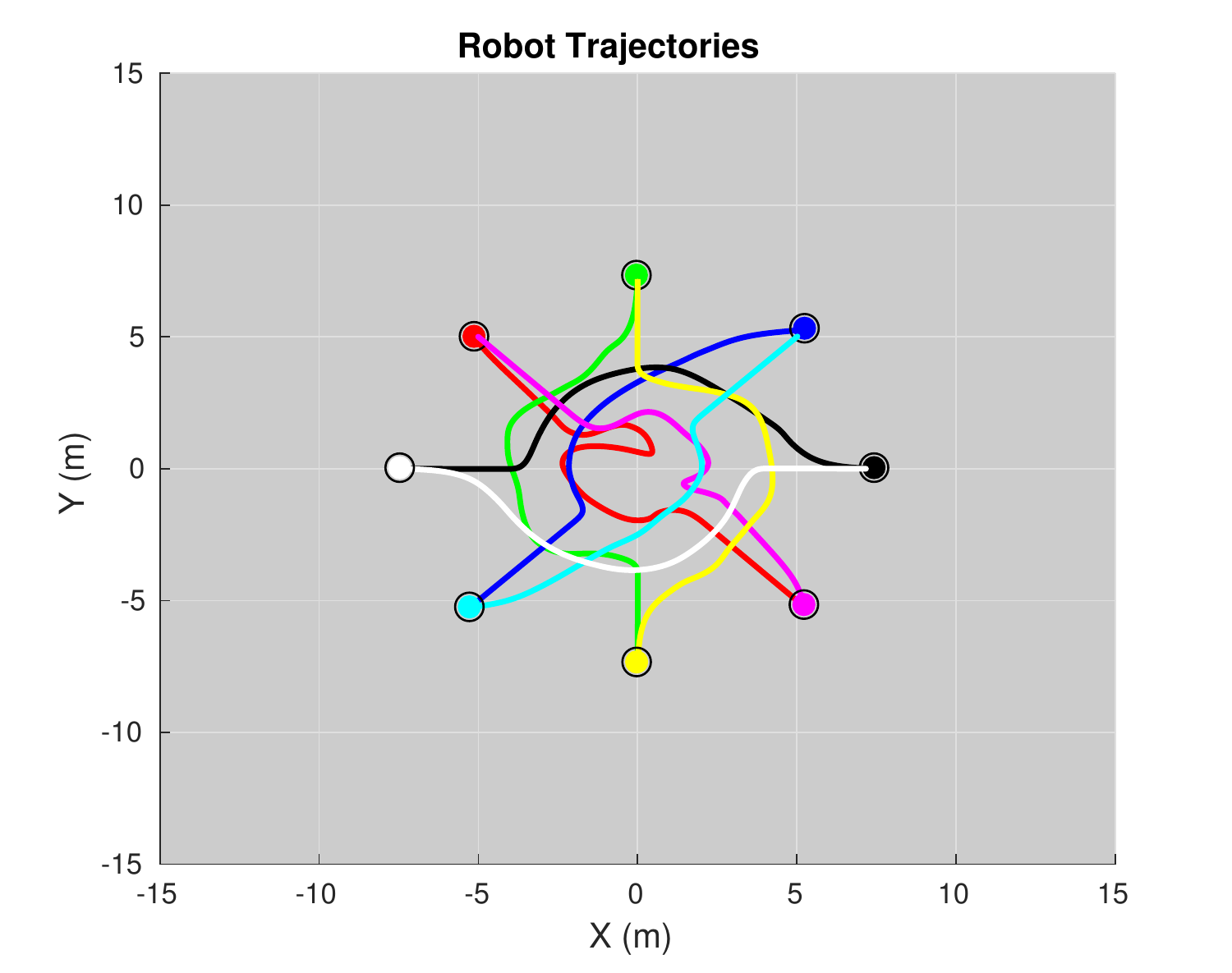} \\
\scriptsize{(a) Approach angle method}  & \hspace*{-28pt} \scriptsize{(b) Tangential band method}
\end{tabular}
\caption{Robot trajectories during antipodal position swapping.}
\label{antipodal_approachangle_tangband}
\end{table}

%

%
\section{Conclusions and Future Work}
\label{sec:conc} 
In this work, we successfully address the problem of obstacle avoidance in the context of decentralized  multi-robot target tracking. Our algorithm uses convex optimization to find collision free motion plans for each robot. We convexify the obstacle avoidance constraints by pre-computing the potential field forces for a horizon and using them as external force inputs in optimization. We show that non-linear dependencies could be converted into such external forces. We validate three methods to avoid field local minima by embedding external forces into the convex optimization. We showcase the efficacy of our approach through gazebo+ROS simulations for various scenarios. Future work involves physically validating the proposed methodology using multiple real robots on different robot platforms (aerial and ground).

\section*{Acknowledgments}
The authors would like to thank Eric Price and Prof. Andreas Zell for their valuable advice during the course of this work.	
%
%

\end{document}